\documentclass[letterpaper, 10 pt, conference]{ieeeconf}  

\usepackage{epsfig}
\usepackage{graphicx}
\usepackage{amsmath}
\usepackage{amssymb}
\usepackage{multirow}
\usepackage{xcolor, soul}
\usepackage{hyperref}
\sethlcolor{yellow}
\usepackage{subcaption}
\usepackage{mathabx}

\IEEEoverridecommandlockouts                              

\newcommand{\systemname}{\textsc{FarSec}}
\newcommand{\etal}{\textit{et al}. }

\title{\LARGE \bf
\systemname: A Reproducible Framework\\ for Automatic Real-Time Vehicle Speed Estimation\\ Using Traffic Cameras
} 

\author{Lucas Liebe$^{\ast}$, Franz Sauerwald$^{\ast}$, Sylwester Sawicki$^{\ast}$, Matthias Schneider$^{\ast}$, Leo Schuhmann$^{\ast}$, Tolga Buz$^{\ast}$, \\ Paul Boes$^{\mathsection}$, Ahmad Ahmadov$^{\mathparagraph}$, Gerard de Melo$^{\ast \blackdiamond}$%
\thanks{$^{\ast}$These authors are affiliated with Hasso Plattner Institute, Prof.-Dr.-Helmert-Str. 2-3, 14482 Potsdam, Germany
       {\tt\small [firstname].[lastname]@student.hpi.de}, {\tt\small \{tolga.buz, gerard.demelo\}@hpi.de}}%
\thanks{$^{\mathsection}$Paul Boes is affiliated with Porsche Digital, Berlin, Germany
        {\tt\small paul.boes@porsche.digital}}%
\thanks{$^{\mathparagraph}$Ahmad Ahmadov is affiliated with Porsche AG, Ludwigsburg, Germany
        {\tt\small ahmad.ahmadov@porsche.de}}%
\thanks{$^{\blackdiamond}$ Gerard de Melo is additionally affiliated with the University of Potsdam, Potsdam, Germany}
}

\begin{document}

\maketitle
\thispagestyle{empty}
\pagestyle{empty}

\begin{abstract}

Estimating the speed of vehicles using traffic cameras is a crucial task for traffic surveillance and management, enabling more optimal traffic flow, improved road safety, and lower environmental impact.
Transportation-dependent systems, such as for navigation and logistics, have great potential to benefit from reliable speed estimation.
While there is prior research in this area reporting competitive accuracy levels, their solutions lack reproducibility and robustness across different datasets.
To address this, we provide a novel framework for automatic real-time vehicle speed calculation, which copes with more diverse data from publicly available traffic cameras to achieve greater robustness. 
Our model employs novel techniques to estimate the length of road segments via depth map prediction. 
Additionally, our framework is capable of handling realistic conditions such as camera movements and different video stream inputs automatically. 
We compare our model to three well-known models in the field using their benchmark datasets.
While our model does not set a new state of the art regarding prediction performance, the results are competitive on realistic CCTV videos. 
At the same time, our end-to-end pipeline offers more consistent results, an easier implementation, and better compatibility.
Its modular structure facilitates reproducibility and future improvements. 
The source code for this project is available on GitHub\footnote{\url{https://github.com/porscheofficial/speed-estimation-traffic-monitoring}}.
\end{abstract}


\section{Introduction}

Traffic speed estimation from monocular video data is a challenge that combines multiple computer vision tasks and has numerous applications in real-world scenarios. 
For instance, traffic management systems aim to monitor traffic flow to ensure smooth traffic, prevent congestion, and improve road safety. 
Surveillance systems need to monitor traffic to detect and prevent accidents, identify traffic violations, and enhance road security. 
In the near future, autonomous driving systems may rely on real-time traffic speed estimations to make decisions regarding vehicle speed, lane position adjustments, and routing.

Traditionally, traffic speed is measured using specialized radar sensors such as LIDAR, or inductive loop systems. 
Some navigation services draw on GPS data from mobile phones and vehicles for this purpose \cite{traffic_speed_est_1}. Despite being reasonably accurate, 
these techniques have major drawbacks: They require expensive hardware, a time-consuming setup and calibration, human interaction, and are not available in all locations and circumstances. 
Meanwhile, monocular traffic cameras are installed on a large number of roads already and traffic speed estimation based on their video data has the potential to offer a complementary source of speed information that is cost-effective and flexible. 
However, video-based methods face several challenges, including the need for accurate object tracking, camera calibration, and efficient inference.
Previous research has addressed the problem of traffic speed estimation with remarkable accuracy -- but achieving real-time performance and end-to-end generalization to new datasets remains a challenging task. 
The most difficult aspect of this process is obtaining a calibration without knowledge of each camera's \emph{intrinsics} (i.e., the camera's optical center and focal length) and \emph{extrinsics} (i.e., its location relative to the scene).

This paper presents \systemname, a novel end-to-end pipeline for calculating vehicle speed from traffic cameras with a more flexible, automatic, and real-time approach. 
The main technical novelty is to obtain more robust estimates by
achieving segment length estimation via depth mapping.
We can then determine the length of the road segment based on the average length of the cars, as detected using a deep convolutional network, and estimate street-specific angles using the center points of tracked vehicles. While this does not obtain new state-of-the-art results on standard benchmarks, it leads to more robust results on real-world camera feeds. 
The overall key contribution is a holistic pipeline that can robustly handle diverse visual traffic camera inputs and adapt to camera movements by automatically re-calibrating. 
We further improve the compatibility and flexibility by handling different stream formats and FPS (frames-per-second) rates, devising a robust vehicle tracking methodology, and eventually computing speed estimates based on a sliding window, which represent the average speeds at which vehicles are travelling on the relevant section of road.
Our contribution constitutes a significant step towards addressing the challenging problem of traffic speed estimation from monocular video data, and provides a versatile and flexible solution that can be used in a wide range of applications by offering an easy-to-deploy setup.

\section{Related Work}

Video-based traffic camera speed estimation and the individual parts of such systems have been investigated from several different angles.
The most relevant research areas for this task are camera calibration, vehicle detection, and vehicle speed estimation.

\paragraph{Camera Calibration}
Numerous studies agree that camera calibration is the key challenge for speed estimation from traffic cameras \cite{compr_brno, robo_auto_mono, autocalib}. 
Common assumptions of existing calibration techniques include \emph{zero pixel skew}, \emph{centered camera principal point}, \emph{plain road}, \emph{straight road segments}, and \emph{constant vehicle speed (no breaking or accelerating)} \cite{compr_brno}. 
A critical quantity for determining speed from video is the amount of meters per pixel in the specific recording, which greatly depends on the video resolution and the orientation of the camera \cite{autocalib}.
Calibration is achieved either manually \cite{zivkovic2004improved} or automatically \cite{dubska2014fully}. 
Our work focuses on automatic approaches, since only these are practical and robust for large-scale real-world deployments.
A precise but complex strategy is to rely on vehicle pose detection \cite{robo_auto_mono, sanchez2020simple}, while some simpler approaches work with vanishing points \cite{orghidan2012camera} or license plates \cite{filipiak2016nsga}. 
For the final mapping of pixel-to-meter information, it is necessary to know the dimensions of pertinent features in the real world, which can be estimated based on the road \cite{traffic_speed_est_1, gerat2017vehicle, czapla2017vehicle}, the vehicles and their properties \cite{robo_auto_mono, czajewski2010vision}, or the lane markings \cite{rao2015real}.

\paragraph{Vehicle Detection}
It is crucial to first identify and then track a moving object in order to determine its speed in a video stream. 
Fernández Llorca \etal survey different vision-based tracking systems and distinguish several methods for tracking a vehicle \cite{survey}.
One approach is the feature-based technique, which records specific vehicle features and follows them throughout the video \cite{czapla2017vehicle, kumar2014vehicle}. 
Pre-processing the video through background subtraction can simplify this task \cite{mohammadi2010vehicle}. 
Another method involves calculating the centroid of the bounding box and tracking it instead of the entire vehicle \cite{rao2015real, sina2013vehicle}. 
The most widely used strategy is to consider entire bounding boxes to account for the entire vehicle's area \cite{gerat2017vehicle, grammatikopoulos2005automatic, liu2020vision}. 
Since cars generally have license plates, another set of approaches focuses on tracking these \cite{czajewski2010vision, llorca2016two}. 
Aside from these traditional methods, machine learning approaches have become increasingly established in this field. 
For example, YOLO
models are widely invoked due to their notable accuracy and efficiency in recognizing bounding boxes of vehicles \cite{liu2020vision, gauttam2020speed, rs11101241}. 
Kocur \etal detect vehicles by using a 3D bounding box and reduce the mean absolute error of the speed measurement estimation by 32\% \cite{kocur2020detection}. 
However, this approach relies heavily on precise camera calibration.

\paragraph{Vehicle Speed Estimation and Traffic Monitoring}
Several different approaches for the final speed estimation exist, as compared in a recent study by Sochor \etal \cite{compr_brno}.
One approach is to use street lines for segment length estimation. However, they argue that these are often non-existent or difficult to recognize.
More accurate and better-performing approaches according to their comparison typically rely on manual adjustments to cater to the specific camera, such as adopting video masks or supplying known distance values \cite{compr_brno}.
One of the best-performing recent approaches \cite{robo_auto_mono} proposes calibration of the traffic camera by minimizing a 3D reprojection error, leveraging general assumptions about the shape and dimensions of cars using a 3D model catalog.
While achieving remarkable results on the BrnoCompSpeed dataset, their approach has several disadvantages:
It requires ground truth video masks provided by prior work \cite{compr_brno}, which also limits their analysis to a single predefined lane. Moreover, they filter out vehicles that fail to get tracked sufficiently well \cite{robo_auto_mono}. Additionally, their Transformer-based architecture is computationally demanding and hence may not be well-suited for real-time deployment.

Previous research presents different approaches to solving parts of the problem, but there is a lack of end-to-end solutions that emphasize compatibility, robustness, and efficiency.
We achieve a holistic solution by devising a robust pipeline that includes provisions for handling multiple stream formats and frame rates, FPS estimation, and camera move detection.
Furthermore, none of the existing approaches utilizes depth maps, which we consider an efficient and powerful approach to solving the main challenge of missing spatial information in monocular video.
Depth maps have of course been studied in other domains \cite{agarwal2023attention}, including self-driving cars \cite{geiger2013vision}, which is close to the use case presented in our work. 

\section{Automatic Speed Estimation with \systemname{}}

\begin{figure}[htp]
    \centering
    \includegraphics[width=0.8\columnwidth]{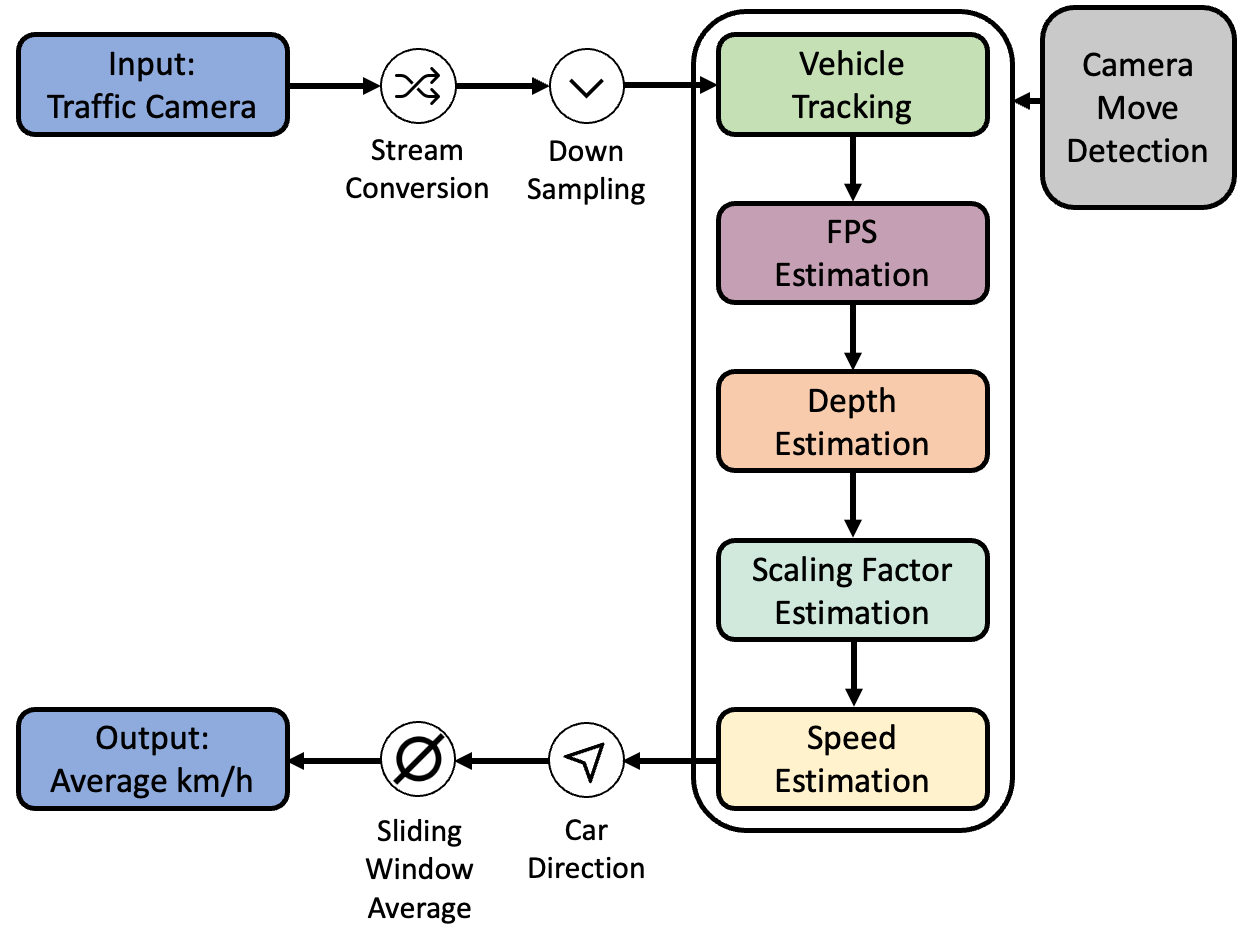}
    \caption{The speed estimation pipeline consists of multiple modules. Camera Move Detection restarts the pipeline in case the traffic camera view switches.}
    \label{fig:pipe}
\end{figure}

\noindent We now introduce the details of \systemname{}, our novel, fully automatic pipeline for real-time sliding window vehicle speed estimation from traffic cameras. The overall approach is illustrated in Figure~\ref{fig:pipe}.

As many traffic cameras stream their footage, our pipeline first converts and downscales the streams to a standard frame rate (of around 30 FPS).
Due to the frequent view changes observed for publicly available traffic camera streams, we incorporated a module to identify these changes and re-calibrate itself accordingly, which operates by detecting average pixel changes.
The tracking of cars is an essential component of this process and is accomplished using a pretrained YOLOv4 model \cite{bochkovskiy2020yolov4}.
Since time is a crucial component in determining a vehicle's speed, the pipeline must read or estimate the number of frames per second that the stream or video includes -- a crucial step that seems to be neglected in a large part of related work. 
We further suggest a novel method for segment length estimation and camera calibration that works with the assumption that automobile dimensions are typical, while deriving their lengths from cumulative tracking of the bounding boxes of the vehicles during their driving path. For relative depth estimation, we incorporate a state-of-the-art depth map model.

Finally, based on the above building blocks, we can generate speed estimates, which are averaged over a sliding window. In the following, we explain each of these components in further detail.

\paragraph{Stream Conversion and Down-Sampling}
Stream conversion and downsampling are important components of the pipeline proposed in this paper to ensure real-time estimations. 
Traffic camera streams can be transmitted via a range of different protocols. 
The most common is the real-time transport protocol (RTP), although some are also provided using the HTTP Live Streaming protocol (HLS). 
In the stream conversion step, HLS or RTP streams are read for further processing.\footnote{The use of Dynamic Adaptive Streaming over HTTP (DASH) is less common, as bitrate-dynamic streaming is not of importance for low-quality CCTV streams.}
We downsample input video streams to a resolution of at most FHD, so as to ensure
that the pipeline can provide real-time estimations.
At higher resolutions, which are rare, but do occur for some video traffic cameras, processing would take substantially longer. The frame rate is capped at $30$ FPS. 
This is particularly important for vehicle tracking. With a lower frame rate, assigning IDs to bounding boxes becomes more challenging, while it works well at $30$ FPS. A higher frame rate, however, incurs a greater processing time without providing valuable extra information. This is explored in detail \hyperref[sec:vehicle-tracking]{further below}.

\paragraph{Camera-Move Detection}
Changes in camera position or alignment can occur for a number of reasons. For instance, some traffic cameras can be remote controlled to cover a larger area. These changes can affect the accuracy of our speed estimations. For a robust solution, it is necessary to detect and re-calibrate the pipeline when such changes occur. Our camera-move detection analyzes the average number of pixels changed between frames. If a spike is encountered, re-calibration of the pipeline is triggered. The threshold is calculated using a rolling average of the last 100 frames and a fixed offset, which can be configured. Due to the rolling average, the detection works reliably with wide- and narrow-view cameras and is robust against small inconsistencies, such as sensor noise, air pollution, or insects flying through the field of view.

\paragraph{Vehicle Tracking}
\label{sec:vehicle-tracking}
Tracking vehicles precisely and consistently across frames is a crucial step for vehicle speed estimation. 
In a first step, cars in a frame of the video are detected.
Here, any approach that yields bounding boxes for each car can be used.
We saw the best performance using a YOLOv4 model pretrained\footnote{Pretrained weights are available at \url{https://github.com/AlexeyAB/darknet\#pre-trained-models}} on the COCO dataset \cite{bochkovskiy2020yolov4}.
The pretrained model accepts images with three channels and a width and height of $608$ pixels. Outputs from the YOLOv4 model are filtered to retain only the \emph{car} class.

Then, each car needs to be assigned an ID and tracked between frames. This is achieved using a nearest neighbour approach with an additional distance threshold of 50 pixels (which is configurable). For each bounding box from the previous frame, the closest bounding box (Euclidean distance) is matched, but only if the distance is below the threshold. In future work, an additional vehicle trajectory estimation could be included to improve matching precision -- this would enable improved handling of video streams with low FPS counts. Assuming a birds-eye perspective on a road with a lane width of $l$ and a camera with $f$ FPS, the maximum trackable speed $s$ with the current approach is
\begin{equation}
s = l \times f.
\end{equation}
In each frame, the car can move at most $l$ meters; otherwise, the matching algorithm may confuse it with an adjacent car.
If the camera has a typical perspective on the road with a vanishing point, the maximum trackable speed is even higher, as the pixel distance of a movement towards the camera, as observed for a car while it is driving, is lower. Thus, the above equation yields a lower bound of the maximum trackable speed.

Using reasonable assumptions, such as a lane width of $3.5 \text{m}$ and a frame rate of $15$ FPS, the maximum trackable speed in a worst case scenario is $52.5 \frac{\text{m}}{\text{s}} \mathrel{\widehat{=}} 189 \frac{\text{km}}{\text{h}}$.
For this reason, we decided to use the fast and robust approach of matching bounding boxes by the difference in centroids. This obtained strong results with the streams we tested, which had around 25--30 FPS.
In future work, alternative tracking methods like DeepSORT \cite{Wojke2017simple} can be investigated.

\paragraph{FPS Estimation}
To calculate the speed from the estimated distances, the absolute time between frames is required. Typically, videos are accompanied by metadata, which includes the number of frames per second (FPS), from which the time between frames can be derived. For traffic camera live streams, such metadata, however, is often missing, making it necessary to estimate the FPS ourselves. We assume that the live stream provides evenly paced frames, i.e., the time between frames is consistent. The FPS estimation module measures the time between frames for a number of frames that can be configured, whenever no metadata is available. With more frames taken into account, latency and jitter (variance in latency) only have a negligible effect on our estimation.

\begin{figure}[htp]
    \centering
    \vspace{0.2cm}
    \includegraphics[width=8cm]{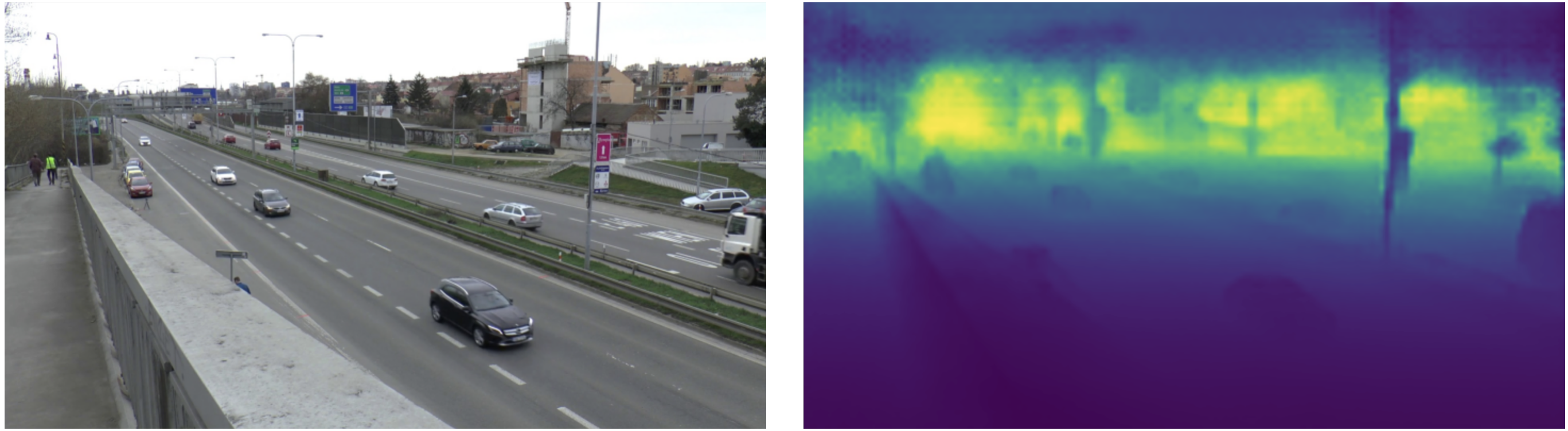}
    \caption{The first step for the segment length estimation is achieved via the Pixelformer depth estimation model.
    Left: one frame from the BRNO dataset, right: the generated depth map.}
    \label{fig:depthmodel}
    \vspace{-0.2cm}
\end{figure}

\paragraph{Depth Estimation}
Depth models are able to infer fairly accurate relative depth estimates. Our approach adopts the Pixelformer \cite{agarwal2023attention} depth estimation model, pretrained on the KITTI dataset \cite{geiger2013vision}. Depth estimates with this model prove accurate despite the training data consisting only of first-person like car-view footage. We evaluated the accuracy by measuring the pixel distance in still frames from videos of the BrnoCompSpeed dataset, which included road markings at a fixed distance. The relative depth output from the model is used to retrieve distances between bounding boxes. An example of the depth model's estimates is given in Figure~\ref{fig:depthmodel}.

As the inference step of the depth estimation model is comparably slow, we calculate a depth map for only a certain number of frames and consider the obtained average relative distances for subsequent frames. However, the relative distance alone does not suffice to determine the vehicle speed, as manual calibration would be needed to obtain the corresponding absolute distance. In \systemname{}, this is instead achieved by calibrating the cameras with a novel approach, which yields a scaling factor that can be applied to the relative distance estimations.

\paragraph{Scaling Factor Estimation}
We use a simple pinhole camera model to translate points of an image $I$ and their associated estimated depths into world points. In particular, for given intrinsic parameters $f$ (focal length), $s_x$, $s_y$ (aspect ratio scaling in $x$ and $y$ direction, respectively) and $u_0$, $v_0$ (principal points), let $p \equiv [u,v,1]$ denote an image point in Cartesian coordinates in the camera coordinate system and let $p \equiv (r_{u,v}, \theta_{u,v}, \phi_{u,v})$ denote the same point in spherical coordinates (still in the camera reference system and using the ISO convention). Then, given a depth map $M_I$ of the image such that $M_I(u,v)$ is the estimated depth in meters of point $p$, we can obtain the corresponding world point $P$ in spherical coordinates (and still in the camera coordinate system) as 
\[
P \equiv (M_I(u,v),~~ \theta_{u,v} + \pi,~~ \phi_{u,v} + \pi).
\]

Unfortunately, models for monocular depth estimation are generally not able to provide depth estimates only up to a scaling factor, meaning that we generally just have $d(u,v) = s_M M_I(u,v)$, where $d(u,v)$ is the true depth of the point $p$ and $s_M$ is a constant (or at least modelled to be one in this work) independent of the image $I$. In general, $s_M$ is unknown and needs to be estimated on top of the camera intrinsics. To do so, we use the following simple least squares estimator: Let $D_M(p,p') := \|p - p' \|_2$ denote the estimated Euclidean distance between two image points (potentially across different images $I$, $I'$ and with the $2$-norm evaluated on the Cartesian representation of the points) based on the depth model $M$. It is easy to show that the \emph{true} distance between the points $D(p,p')$ satisfies $D(p,p')= s_M D_M(p,p')$. Hence, we can use a sequence of pairs of image points $\mathcal{P} = ((p_i, p_i'))_i$ with associated true world point distances $L = (l_i)_i$ to estimate $s_M$. 
We denote as $D_M(\mathcal{P})$ the vector of estimated distances from the pairs $\mathcal{P}$ without the scaling factor.
We can then solve for 
\begin{equation}
\min_{\hat{s}_M \in \mathbb{R}} \|L - \hat{s}_M D_M(\mathcal{P})\|_2,
\end{equation}
which is just a least squares problem and has the well-known solution
\begin{equation}
\hat{s}_M = \frac{D_M(\mathcal{P})^\intercal L}{\|D_M(\mathcal{P})\|^2}.
\end{equation}
In practice, we use average car lengths as ground truth distances, where a combination of the 2D tracking boxes from the car tracking step is combined with the projection of the box centroids over time. However, other, for example country-specific, characteristics could be used.

\paragraph{Speed Estimation}
\systemname{} uses the scaling factor estimation from the initial depth map images to calculate the distance a vehicle travels.
After collecting the bounding boxes and assigning IDs, we track the centroids of the bounding boxes for each vehicle through the video and derive a centroid line.
We then calculate the intersections of each vehicle's bounding box with this centroid line, which defines multiple ground truth events.
The pairwise intersection points serve as input for the least squares estimator from above. 
This estimator provides the Euclidean distance between the two intersection points under the assumption of the mean vehicle length (including bounding box padding) of six meters. 
Additionally, each car is assigned a traffic direction, derived from the sign of the difference of the $y$ coordinate (image height) for a car. This is important, as different traffic directions have different road segments assigned, which need separate speed information.

To calculate an average speed for one direction, all cars for that direction are then combined in a rolling average for the last $p=60$ seconds. Let $V = (v_1, v_2, ..., v_n)$ be a sequence of average speeds for all $n$ cars and $T = (t_1, t_2, ..., t_n)$ the corresponding sequence of timestamps (in seconds) when each car was last seen. We compute the following:
\begin{equation}
    \begin{split}
    V_t &= \{ v_i \mid \exists i \in \mathbb{N}: 1 \leq i \leq n \,\land\, t_i > t - p\} \\
    v^*(t) &= \frac{\sum_{i=0}^{|V_t|} V_t}{|V_t|} \\
    \end{split}
\end{equation}
This computation is done at any time $t$ for which an output is desired. The parameter $p$ can be configured depending on the particular use case.

\paragraph{Model Design \& Applicability}

\systemname{} offers an automatic end-to-end pipeline, covering all steps necessary to get from input files or streams to a machine-readable speed estimation output, without the need for manual intervention.
Existing solutions either use manual calibration to adapt to new datasets \cite{survey, kocur2020detection} or require costly preprocessing of the input data to apply masking \cite{dubska2014fully}, calculate vanishing points \cite{compr_brno}, etc. 
A few other solutions do offer automatic calibration, but lack an easy integration and are computationally expensive \cite{robo_auto_mono}.
Due to its automatic calibration and its robust design that emphasizes compatibility with different camera setups, our approach is scalable and easily applied to new and different datasets. 
To ensure that \systemname{} is compatible with a variety of traffic cameras and easily applied to other datasets, our pipeline has been developed to be compatible with camera feeds from OpenTrafficCamMap \cite{welch2020open}, which offers a total of 32,221 cameras across the United States. 
Unfortunately, we are not allowed to share this dataset due to license restrictions.
While the motivation for our research was to offer a framework for speed estimation, simpler use cases such as counting cars per direction can also easily be covered by our pipeline.

\section{Evaluation}

\subsection{Experimental Setup}
\label{sec:experimental-setup}

\paragraph{Datasets}\label{para:datasets}
We draw on two different datasets for the evaluation of our traffic speed estimation pipeline \systemname:
The BrnoCompSpeed \cite{compr_brno} and the CCTV dataset published by Revaud \etal \cite{robo_auto_mono}.

BrnoCompSpeed \cite{compr_brno} consists of six sessions, which capture street scenarios in six different locations.
Each session includes three videos, showing the same location from three different vantage points (left, right, center) -- leading to a total of 18 videos.
The ground truth speed for each vehicle was measured using LIDAR.
The videos have a rather high resolution of 1,920 × 1,080 pixels compared to regular traffic camera datasets, and all videos were captured under neutral weather conditions and during daylight.
The data thus excludes settings such as precipitation and night scenes, and does not reflect the full variety of real-world scenarios.

Revaud \etal provide a similar critique of BrnoCompSpeed in their work \cite{robo_auto_mono} and published an alternative CCTV dataset, which we use for additional benchmarking.
Their dataset comprises 40 videos recorded from public CCTV cameras in South Korea, each approximately two minutes long. Each video, however, only contains a single vehicle with ground truth speed, visible only for a short period within the video.
The ground truth was fixed by manually annotating cars in the video footage, using GPS tracking devices and driving through the respective scene \cite{robo_auto_mono}.

\begin{figure}[htp]
    \vspace{0.2cm}
    \centering
    \includegraphics[width=0.5\columnwidth]{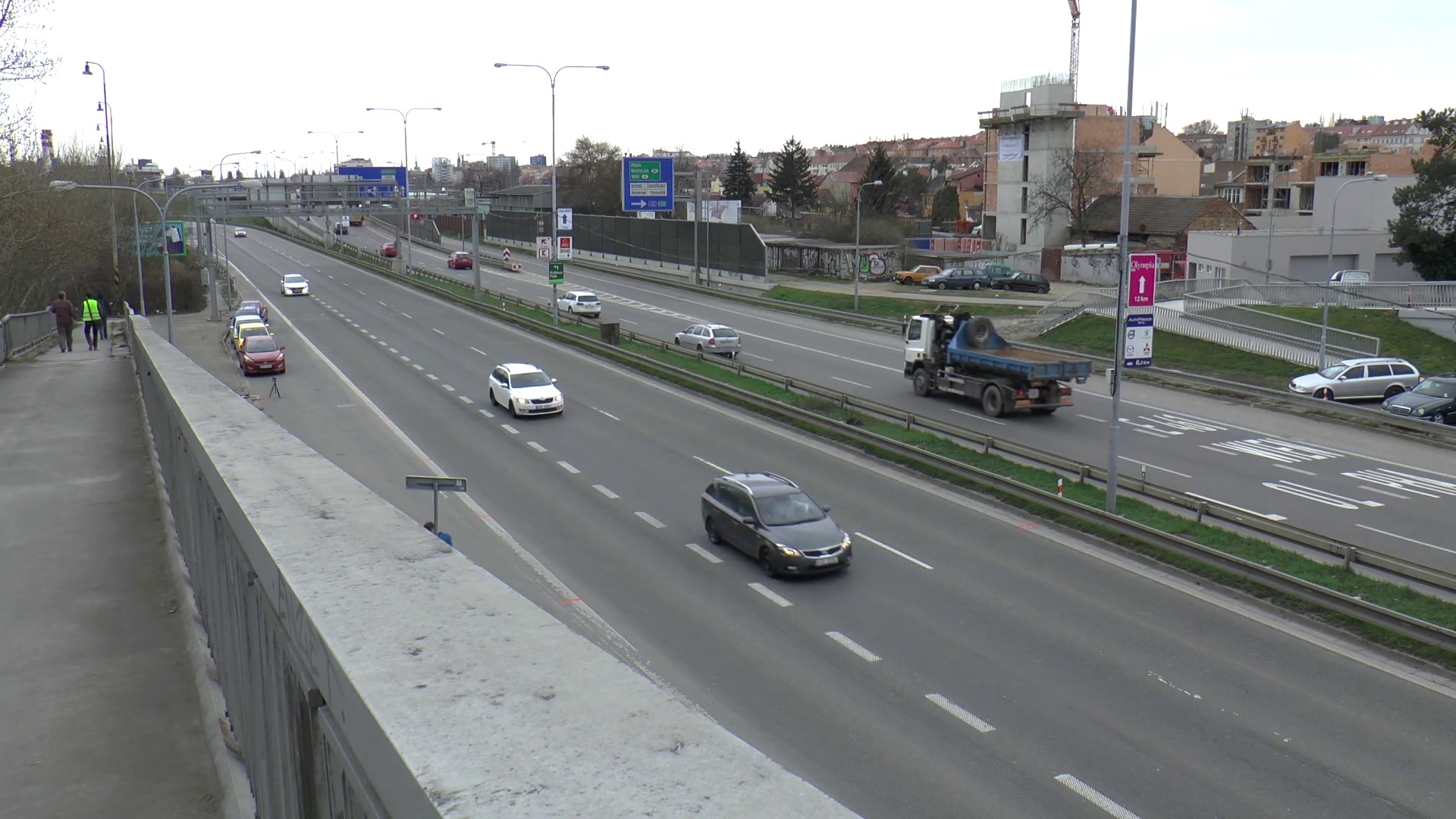}
    \includegraphics[width=0.5\columnwidth]{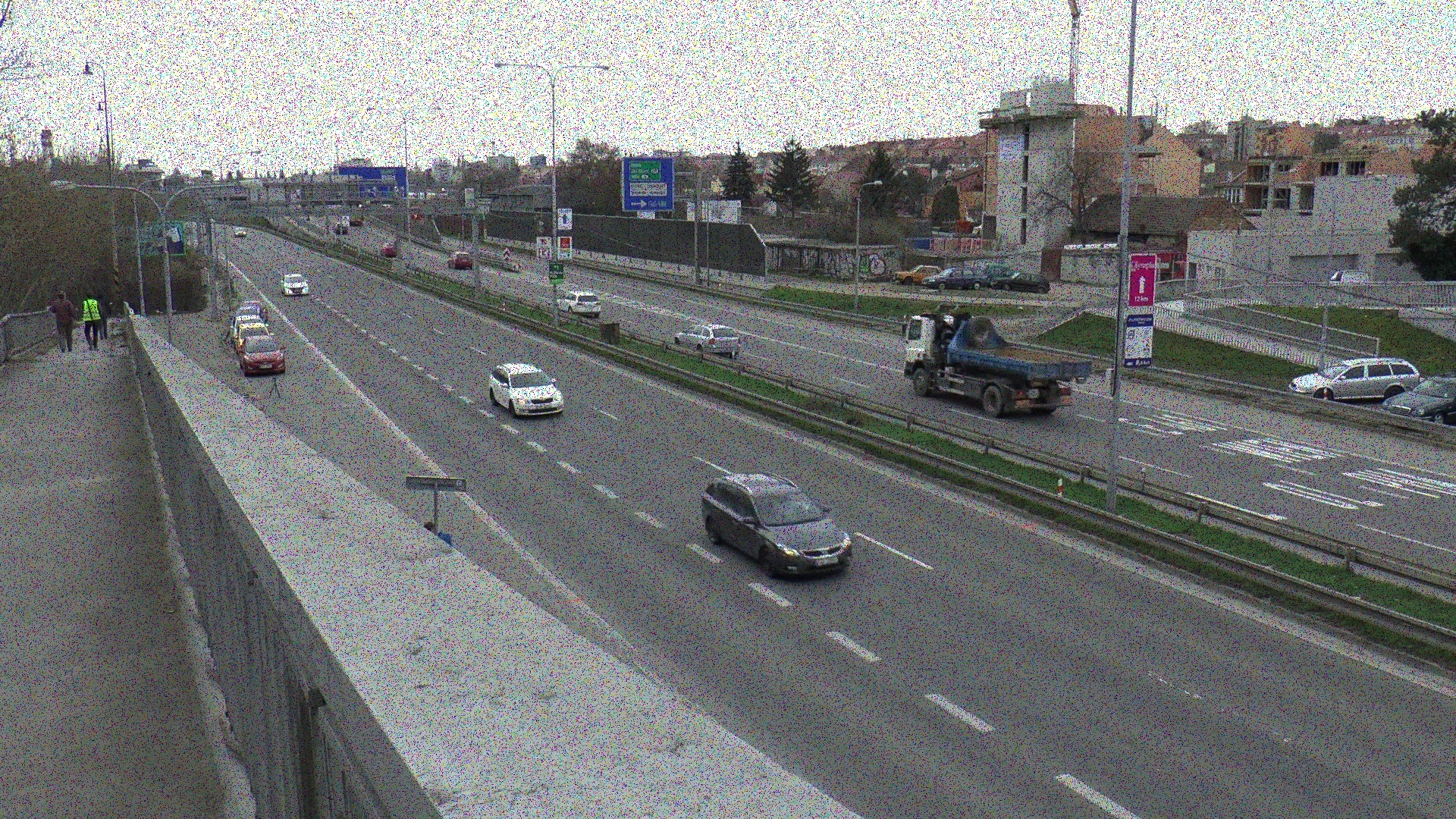}\hfill
    \includegraphics[width=0.5\columnwidth]{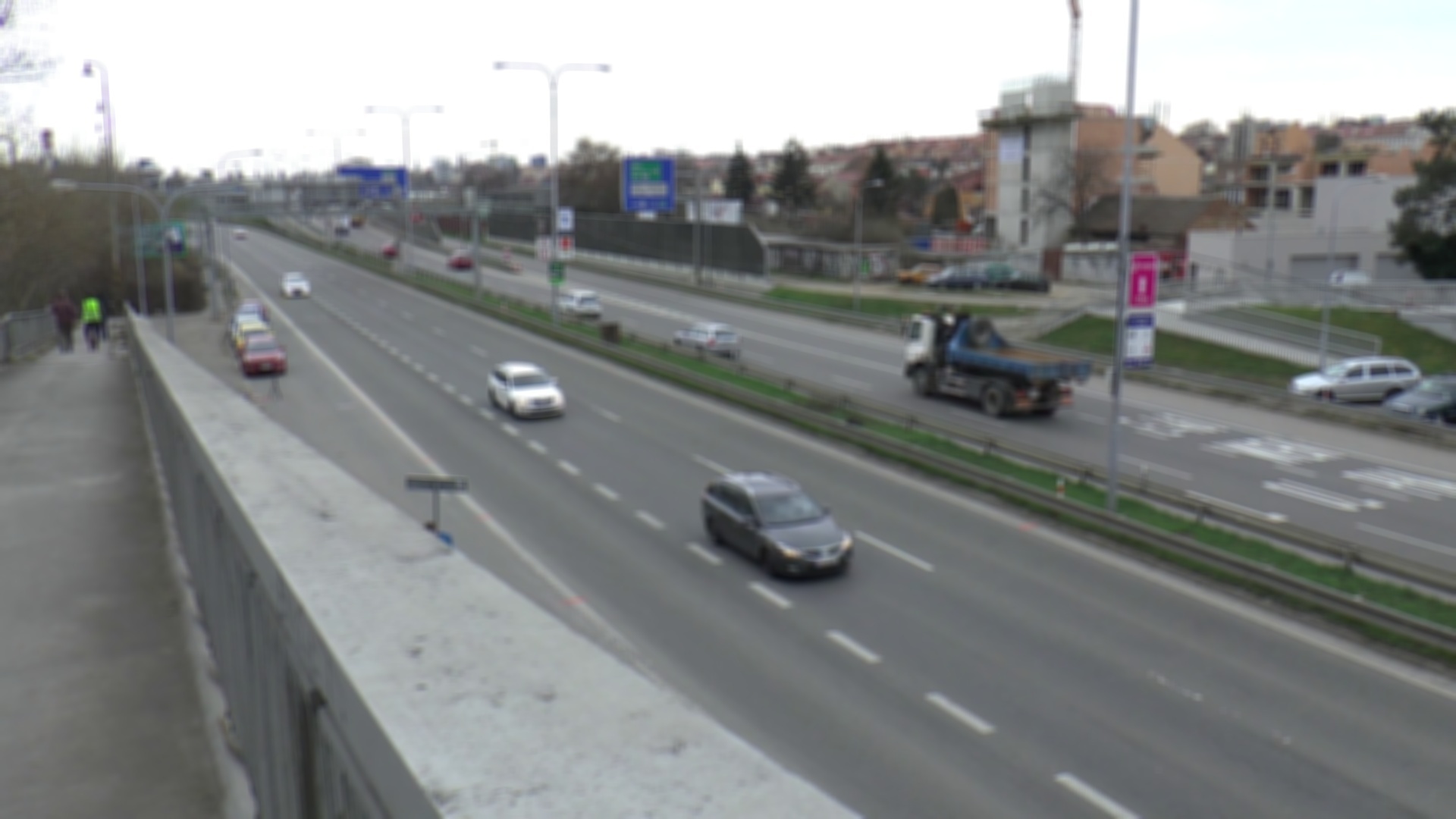}
    \caption{Example of augmentation step for frame 200 of the BrnoCompSpeed dataset (session six left) \cite{compr_brno}: original (top left), augmented with noise (top right), augmented with blur (bottom).}
    \label{fig:brno_hist_abs}
    \vspace{-0.2cm}
\end{figure}

Unfortunately, both benchmark datasets are very limited regarding the number of ground truth samples and videos in total.
In order to increase the amount of evaluation data, we additionally augment the BrnoCompSpeed dataset in two different ways to further test our pipeline -- by applying blurring and by applying a noise filter.
Figure \ref{fig:brno_hist_abs} shows one frame of the dataset, in its original form as well as the corresponding blurred and noisy versions.
The blurring is achieved by convolving an image with a normalized box kernel of size $10 \times 10$ as a filter. The central element is replaced by the average of all pixels in the kernel area. 
Noise is added by randomly replacing 10\% of all pixels with a fixed value (1 in our case).
This augmentation allows the full-HD dataset to approximate the image quality of a CCTV camera, while also enabling an evaluation of  the robustness of our pipeline.


\paragraph{Baseline Models}
We select Learned+RANSAC \cite{robo_auto_mono}, Full\-ACC \cite{dubska2014fully}, and OptScaleVP2 \cite{compr_brno} as benchmark reference models, due to their presence in related research, availability of benchmark results, and their similarity to our approach, being (mostly) automatic models for the same task (although OptScaleVP2 requires a semi-manual calibration by entering road distance measurements for the analyzed videos).
Sochor \etal \cite{compr_brno} additionally propose OptCalibVP2, which they claim performs better with significantly lower error rates than FullACC and OptScaleVP2.
However, OptCalibVP2 utilizes ground truth values for finding an optimal calibration using grid search \cite{compr_brno}, which is why we consider it overfit for the BrnoCompSpeed benchmark and not representative of a model that could be applied to new traffic camera datasets.
Furthermore, it should be noted that it was not possible for us to reproduce any one of these models -- FullACC \cite{dubska2014fully} and OptScaleVP2 \cite{compr_brno} do not provide the source code for their solutions, while we found the repository provided by Revaud \etal \cite{robo_auto_mono} to be incomplete (see \emph{Model Reproducibility} below for further details).

\subsection{Experimental Results}

\paragraph{BrnoCompSpeed Benchmark}

\begin{figure}[htp]
    \centering
    \vspace{0.2cm}
    \includegraphics[width=8cm]{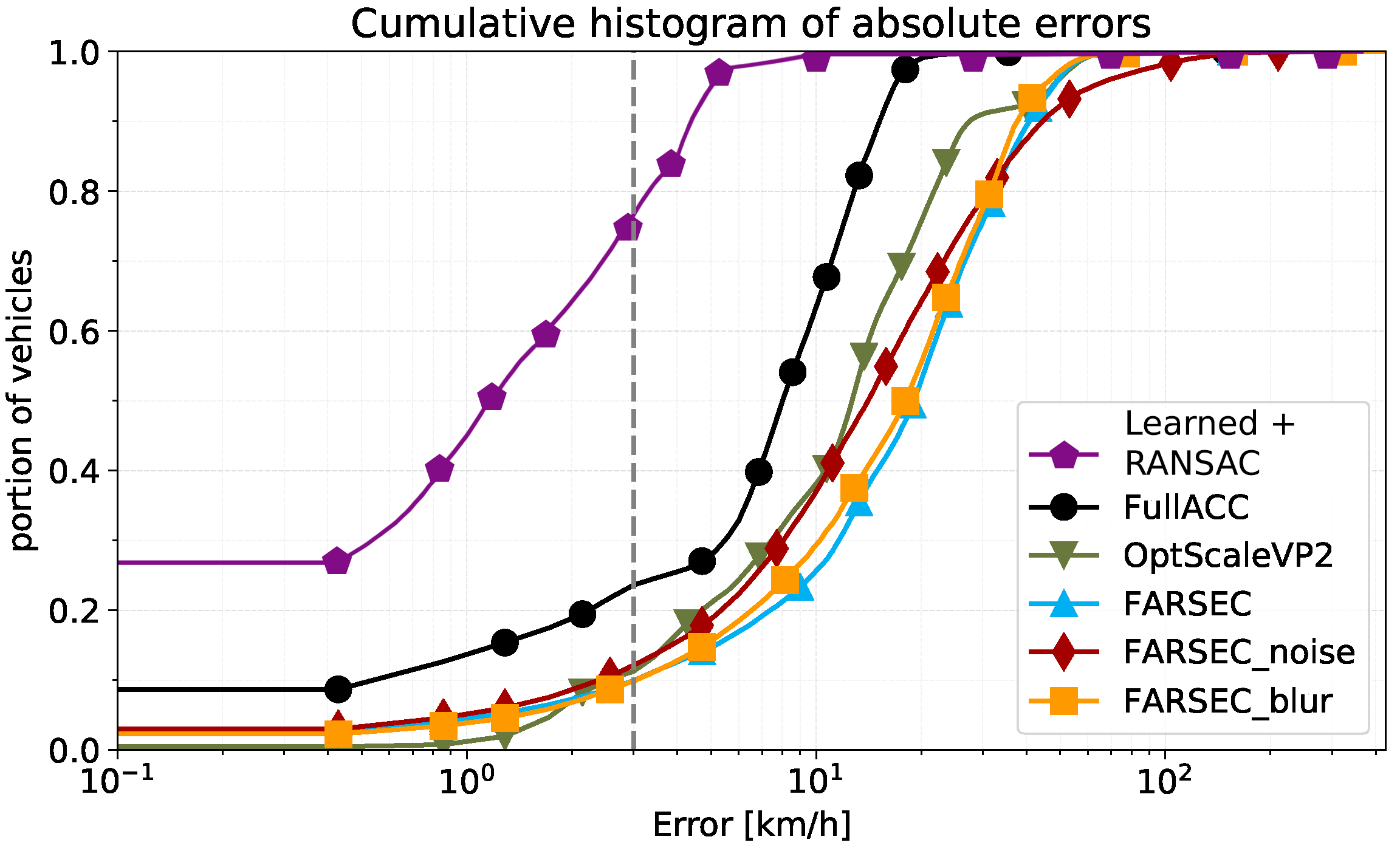}
    \caption{The cumulative absolute error of the different models on BrnoCompSpeed \cite{compr_brno} plotted in a histogram, including our model's benchmarking results for the augmented datasets with noise and a blur filter. The threshold of a 3 km/h error is indicated by the gray vertical dashed line.}
    \label{fig:brno_hist_abs_full}
\end{figure}

The results of our model variants as well as the baseline models (without augmentation) on the BrnoCompSpeed dataset are plotted in Figure \ref{fig:brno_hist_abs_full}. Aggregated results are given in Table \ref{tab:brno_full_results}.
On this benchmark, our model does not quite match the reported results of non-real-time models, which we have not been able to confirm or disprove due to their lack of reproducibility.
Our real-time pipeline has a mean absolute error of 21.24 km/h across all videos, which is higher than for the other approaches, but still reasonable for many use cases.
A detailed review of the performance on individual videos reveals that our pipeline does outperform the non-real-time approaches on some videos.

The results of the runs on augmented data indicate the robustness of our solution.
As can be seen in Figure \ref{fig:brno_hist_abs_full} and Table \ref{tab:brno_full_results}, the pipeline presented in this paper can handle both forms of augmentation well. 
Our model performs best on blurred data, achieving slightly better results than the variant tested on non-augmented videos of BrnoCompSpeed.
The ability to achieve similar performance with blurred or noisy data indicates that our setup ensures robustness for lower-quality datasets and does not exhibit overfitting to the specific scenes of BrnoCompSpeed.
Due to the lack of reproducibility of prior approaches, we are not able to evaluate the other models on the modified BrnoCompSpeed videos. 

\begin{table}
    \caption{Mean error of our pipeline variants (\systemname{}, \systemname{} blur, \systemname{} noise) and the benchmark results of Learned+RANSAC, FullACC, and OptScaleVP2, as reported in their respective publications evaluated on all videos of the BrnoCompSpeed dataset.}
    \label{tab:brno_full_results}
    \vspace{-0.2cm}
    \begin{center}
        \resizebox{\columnwidth}{!}{%
            \begin{tabular}{|l|r|r|r|r|r|}
            \hline
                                                 &           & \multicolumn{2}{c|}{Abs error (km/h)} & \multicolumn{2}{c|}{Rel error (\%)} \\
                                                 &   support &     mean  & median &  mean  &  median \\
            \hline\hline
            \multicolumn{6}{|l|}{\emph{Non-Real-Time}} \\
            Learned+RANSAC \cite{robo_auto_mono}   &    19,780 &      2.15 &   1.60 &  2.65 &  2.07 \\ 
            FullACC  \cite{dubska2014fully}      &    18,398 &      8.59 &   8.45 & 10.89 & 11.41  \\
            OptScaleVP2  \cite{compr_brno}       &    18,398 &     15.66 &  13.09 & 19.83 & 17.51\\
            \hline
            \multicolumn{6}{|l|}{\emph{Real-Time}} \\
            \textbf{\systemname{}}                        &    \textbf{15,528} &     \textbf{21.24} &  \textbf{19.46} & \textbf{26.34} & \textbf{23.34}  \\
            \textbf{\systemname{} blur}                  &    \textbf{15,824} &     \textbf{20.16} &  \textbf{18.45} & \textbf{25.12} & \textbf{22.44}  \\
            \systemname{} noise                 &    14,330 &     21.28 &  14.53 & 26.79 & 18.14  \\
            \hline
            \end{tabular}%
        }
    \end{center}
    \vspace{-0.2cm}
\end{table}

\paragraph{CCTV Dataset Benchmark}

\begin{figure}[htp]
    \centering
    \includegraphics[width=0.8\columnwidth]{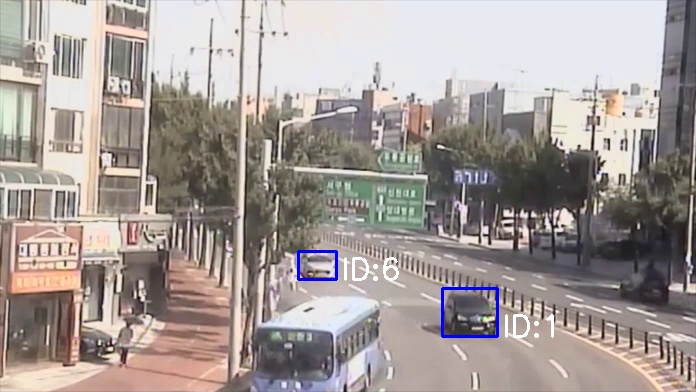}
    \caption{Example of a challenging scene for detection and tracking: the ground truth vehicle can be seen in dark color, without a bounding box, driving away from the CCTV camera on the outer right edge of the frame.}
    \label{fig:cctv_ground_truth_not_detected}
\end{figure}

The benchmarking on the more realistic CCTV dataset \cite{robo_auto_mono} shows more relevant results.
We find that our pipeline has a mean absolute error of 12.22 km/h (see Table \ref{tab:iccv_results}), which is a competitive result.
Note that different versions of this data have been used for evaluation. 
We understand based on our exchange with them that Revaud \etal could not utilize all the videos of the dataset for their benchmarking. 
For \systemname{}, the evaluation is computed based on a subset of 21 videos:
One video is omitted (number ten), since the authors notified us that its values are incorrect.
In 18 videos, the YOLO model is not able to detect the only ground truth vehicle -- so those videos cannot be evaluated for our pipeline. 
This also shows that, for further research datasets, it would be helpful if ground truth speeds for all passing vehicles could be provided.
The task setup for the current data involves targeting a particular vehicle, while \systemname{} instead focuses on \emph{average} speeds of detected vehicles. 
The subpar YOLO vehicle detection on the CCTV data is an artifact of challenging scenery, in which the detection of individual vehicles is highly non-trivial (see Figure \ref{fig:cctv_ground_truth_not_detected}). 
Also, vehicles appearing in just a few frames are disregarded by \systemname{} in favor of vehicles that can be more reliably tracked.
If tracking speeds of individual vehicles is desired for a particular use case, one could switch to an alternative tracking model that picks up vehicles appearing in just a few frames.
Due to the modular structure of our pipeline, individual components can easily be swapped out.

Overall, these results suggest that our pipeline can handle real-world scenarios, and perform even better there than on the BrnoCompSpeed dataset.
The higher relative errors of different models on the CCTV data compared to BrnoCompSpeed stem from the fact that the former focuses on lower-speed urban traffic, so large relative errors are often small in absolute terms.
Regarding the median error, Learned+RANSAC \cite{robo_auto_mono} performs very well -- however, its significant difference between mean and median indicate that it sometimes produces severe outliers. It should be noted that since we were not able to reproduce their results, we use previously reported results  \cite{robo_auto_mono}. 
Interestingly, the performance of the Full\-ACC++(*) approach, introduced by Dubsk{\'a} \etal \cite{dubska2014automatic} \cite{dubska2014fully} and reimplemented by Revaud \etal \cite{robo_auto_mono}, is far off from the other approaches compared in Table \ref{tab:iccv_results}.
In summary, our results show that \systemname{} performs well across diverse datasets and achieves strong accuracy levels also on more realistic CCTV footage.
In future work, a comparison of the inference time of our model could be interesting -- as \systemname{} is able to handle video streams with 30 frames per second in real-time.


\begin{table}
    \newcommand{\invzero}{\hphantom{0}}
    \vspace{0.2cm}
    \caption{Comparison of the mean error for the speed estimation on the CCTV dataset from Revaud \etal \cite{robo_auto_mono} (FullACC++* is their re-implementation of FullACC \cite{dubska2014fully}). Note that here Learned+RANSAC is considered real-time.}
    \label{tab:iccv_results}
    \vspace{-0.2cm}
    \begin{center}
        \resizebox{\columnwidth}{!}{%
            \begin{tabular}{|l|r|r|r|r|}
            \hline
                                        & \multicolumn{2}{c|}{Abs error (km/h)} & \multicolumn{2}{c|}{Rel error (\%)} \\
                                        & Mean & Median                        & Mean & Median                     \\
            \hline\hline
            \multicolumn{5}{|l|}{\emph{Non-Real-Time}}  \\
            FullACC++*  \cite{robo_auto_mono}                  &  32.60    & 22.40 & 56.00 & 52.90  \\

            \hline
            \multicolumn{5}{|l|}{\emph{Real-Time}}  \\
            Learned+RANSAC \cite{robo_auto_mono}               &   12.80    & 5.82 &  29.40 & 16.00 \\
            \textbf{\systemname{}}                              & \textbf{12.22} & \textbf{11.19} & \textbf{44.59} & \textbf{36.11}  \\
            \hline
            \end{tabular}%
        }
    \end{center}
    \vspace{-0.2cm}
\end{table}

\paragraph{Model Reproducibility}\label{para:model_rep}

FullACC \cite{dubska2014fully} and OptScaleVP2 \cite{compr_brno} do not provide their source code, and can therefore not easily be reproduced.
The FullACC++ re-implementation of Revaud \etal \cite{robo_auto_mono} (which became possible when the authors of Full\-ACC shared code snippets with them) shows inconsistent results -- while it seems to perform remarkably well on BrnoCompSpeed, its performance is significantly worse on the CCTV dataset.
This inconsistency indicates issues that could be caused by difficulties to re-implement the model, or, if re-implemented correctly, potential overfitting to BrnoCompSpeed, which causes significantly worse predictions on other datasets.
Revaud \etal do not provide a complete end-to-end solution in their repository.
One problem we encounter while investigating their published pipeline is the missing vehicle detection model.
It is required to plug in your own model or use an open source one that they refer to.
Even after setting up a detection model, the calculation of the homographies fails in our tests.
Additionally, we have faced further questions regarding their filtering of tracks and reproducibility of their random sampling procedure.

\paragraph{Further Evaluation Criteria}

In addition to the performance benchmarks, we identify key criteria for traffic speed estimation systems, including code availability, reproducibility, and compatibility with new traffic camera datasets.
Our work with U.S.-based traffic cameras highlights these factors (shown in Table \ref{tab:comparison}).
\systemname{} does not consistently outperform existing models in terms of the mean error, but performs consistently well on real-world inputs. 
The baseline models seem to rely heavily on high-resolution video inputs, which is rarely available in real-world CCTV streams.
Models that only work with standard video formats such as MP4 or AVI are not well-equipped for real-world applicability, since most public traffic cameras provide their videos as a stream. 
In contrast to other models, \systemname{} handles \emph{different stream formats} out-of-the-box.
During exploration of the U.S.\ traffic camera dataset, we discovered that some cameras change their viewpoint regularly, which would cause increased errors in other models due to unsuitable calibration and their pipelines not being designed for handling such exceptions.  
In contrast, \systemname{} automatically re-calibrates itself upon detecting camera movements.
To ensure reproducibility and facilitate model comparisons, we have made our system available as open source at \url{https://github.com/porscheofficial/speed-estimation-traffic-monitoring}.
\systemname{} is built to be as easily adaptable to new camera data as possible -- a property that can best be evaluated in future work, once more benchmark datasets have been published.
Our pipeline's modular end-to-end design easily provides new speed estimations when changing to new video streams.

\begin{table}
    \vspace{0.2cm}
    \caption{Comparison of \systemname{}'s to existing models regarding relevant criteria (A: BrnoCompSpeed benchmark \cite{compr_brno}, B: CCTV benchmark \cite{robo_auto_mono}, both indicate absolute error in km/h; FullACC is re-implemented for B by the authors of prior work \cite{robo_auto_mono})}
    \label{tab:comparison}
    \vspace{-0.2cm}
    \begin{center}
        \resizebox{\columnwidth}{!}{%
            \begin{tabular}{|l|r|r|r|r|}
            \hline
                                    & Learned+                      & FullACC &  OptScaleVP2 & \multirow{2}{*}{\textbf{\systemname{}}}  \\
                                    & RANSAC \cite{robo_auto_mono} &  \cite{dubska2014fully}  & \cite{compr_brno} &      \\
            \hline\hline
            Mean error A            &  2.15 &  ~8.60 & ~15.70 & \textbf{~20.10} \\ 
            Mean error B            &  12.80 &  32.60 \cite{robo_auto_mono} & N/A & \textbf{12.22} \\ 
            Auto calibration        &  yes       &  yes  &  no   & \textbf{yes} \\
            Stream formats          &  no        &   no  & no    & \textbf{yes}  \\
            Move detection          &  no        &   no  & no    & \textbf{yes}  \\
            Code available          &  partially &  no   & no    & \textbf{yes}  \\
            Reproducible            &  no        &   no  & no    & \textbf{yes}  \\
            End-to-end              &  yes       &   no  & no    & \textbf{yes}   \\
            New dataset             &  yes       &   yes & yes   & \textbf{no}   \\
            \hline
            \end{tabular}%
        }
    \end{center}
    \vspace{-0.2cm}
\end{table}

\subsection{Discussion}
\noindent 
Our results show that while our new real-time framework \systemname{} does not establish a new state of the art regarding mean prediction error, the presented results are competitive with non-real-time approaches (some of which require calibration), especially on the more realistic CCTV dataset benchmark. 
In addition, we contribute an end-to-end pipeline that is designed for flexibility and compatibility with real-world datasets, with an emphasis on accessibility and reproducibility.
Our solution is available as an open source repository to facilitate reproducibility.
The results of our model are sufficiently accurate to be used in traffic management systems, e.g., to refine car navigation, improve traffic flow to reduce congestion, and improve road safety. 
This can help in reducing the environmental impact of transportation by increasing efficiency without the need to install new, expensive hardware on many streets around the world.
The modular design of our pipeline offers opportunities for future work to further improve the performance of \systemname{}.
In addition to the two benchmark datasets shown in our evaluation, our system's modules could be evaluated separately, e.g., the vehicle tracking module could be assessed with a dataset such as UA-DETRAC \cite{CVIU-UA-DETRAC}. 
For brevity reasons, we are focusing on the two available end-to-end benchmarks for this work and encourage the research community to further iterate on \systemname{} using our open source repository.

\paragraph{Limitations}
Our available benchmark datasets only allow evaluation in daytime and neutral weather conditions.
Hence, our car tracking and depth map creation strategies may struggle in low-light or low-visibility settings -- however, our modular setup allows replacing these parts with better alternatives in the future, as long as the source videos provide sufficient signal quality. 
Similarly, there is only limited benchmark data showing curved road segments -- nonetheless, our pipeline performed well on a small set of examples of curved roads that we reviewed manually.
Our methodology relies on cars driving on the road segment to calibrate the scaling factor -- if only large trucks or motorbikes are visible in the video stream, the parameter for the average car length may not be correct.

\paragraph{Societal Impact}
Using traffic video information comes with the downside of privacy concerns. Our work relies on streams that are already publicly available.
Such publicly accessible traffic cameras are typically provided at a low resolution, making very hard to recognize faces or people inside a car. As monocular speed estimation techniques are subject to estimation errors, these systems should not be utilized as a substitute for radar traps for policing traffic speed violators.

\section{Conclusion}
\noindent We present an automatic end-to-end pipeline for monocular vehicle speed estimation that manages the different challenges of this task in a novel way. 
It is capable of processing traffic camera streams in real-time and provides new solutions for speed estimation via depth-estimation and scaling factor calculation via aggregation of car trails.
While our approach achieves competitive performance on the more realistic CCTV benchmark, its main benefits lie in its calibration-free setup, modularity and reproducibility.
As additional, larger, realistic benchmark datasets emerge, our results suggest that \systemname{} will continue to perform competitively.
The modular setup allows for further improvements by exchanging specific parts of the pipeline in future work.
As we provide as an easy-to-deploy open-source pipeline with real-time processing, it may serve as a cheap, out-of-the-box vehicle speed estimation approach for general traffic analysis or as input for navigation system models. 
Our source code has been made freely available on \href{https://github.com/porscheofficial/speed-estimation-traffic-monitoring}{GitHub}.



\bibliographystyle{ieeetr}  
\bibliography{references}

\appendix

\section{Detailed Evaluation Results}

Tables \ref{tab:fullacc_results}, \ref{tab:optscale_results}, \ref{tab:farsec_results}, \ref{tab:farsec_blur_results}, \ref{tab:farsec_noise_results}, and \ref{tab:farsec_bg_results} present the evaluation results of FullACC, OptScaleVP2, and \systemname{} (with original, blurred, noisy, and background removed input data), respectively, on the BrnoCompSpeed benchmark dataset.
The results of FullACC and OptScaleVP2 are obtained by utilizing the provided result data of the respective authors \cite{dubska2014fully,compr_brno} and the provided evaluation script of Sochor \etal \cite{compr_brno}.
Unfortunately, this detailed evaluation could not be generated for the Learned+RANSAC model from Revaud \etal \cite{robo_auto_mono} due to its lack of reproducibility (as explained earlier).
Table \ref{tab:all_approaches_best_results} shows the mean (km/h) errors of all evaluated approaches on the BrnoCompSpeed benchmark dataset \cite{compr_brno}. The lowest error is indicated in bold.
While the results of these models appear very strong, with FullACC achieving a mean absolute error below 9 km/h, it is not possible for us to replicate their methods of creating the results.
This is due to their source code not being publicly available.
Table \ref{tab:cctv_full_results} provides further evaluation of \systemname{} on the CCTV dataset \cite{robo_auto_mono}.
It should be noted that the baseline models from related research, Learned+RANSAC, FullACC, and OptScaleVP2, all perform significantly worse on the more difficult CCTV dataset compared to their performance on BrnoCompSpeed, as reported and presented in prior work \cite{robo_auto_mono}, while \systemname{} achieves more consistent results and even performs better on the more difficult dataset.


\begin{table*}
\begin{center}
    \caption{Evaluation results of the FullACC model on the BrnoCompSpeed benchmark}
    \begin{tabular}{|l|r|r|r|r|r|}
    \hline
        video      &     support   &    mean [km/h]   &    median [km/h]   &    95th percentile [km/h]  &     worst [km/h]   \\
    \hline\hline
    s1 left        &                 749 &            11.18 &              11.00 &                   13.70 &              51.44 \\
    s1 center      &                 493 &            10.15 &               9.12 &                   17.81 &              77.84 \\
    s1 right       &                 607 &             6.81 &               6.65 &                   11.22 &              31.53 \\
    s2 left        &               1,030 &            16.24 &              16.14 &                   19.72 &              36.91 \\
    s2 center      &               1,208 &            11.14 &              10.69 &                   14.72 &              33.22 \\
    s2 right       &               1,284 &             0.95 &               0.77 &                    1.45 &              21.03 \\
    s3 left        &                 173 &            14.28 &              13.99 &                   19.75 &              27.55 \\
    s3 center      &                 187 &            13.96 &              13.90 &                   17.58 &              22.23 \\
    s3 right       &                 177 &             7.54 &               7.27 &                   10.85 &              15.57 \\
    s4 left        &               1,026 &            15.88 &              15.82 &                   18.95 &              22.48 \\
    s4 center      &               1,056 &             1.10 &               0.62 &                    2.61 &              27.59 \\
    s4 right       &                 931 &             2.05 &               1.81 &                    3.79 &              25.88 \\
    s5 left        &               1,944 &             6.99 &               7.08 &                    8.87 &              15.53 \\
    s5 center      &               1,932 &             9.64 &               9.11 &                   14.38 &             119.32 \\
    s5 right       &               1,950 &            12.02 &              11.83 &                   19.39 &              68.10 \\
    s6 left        &               1,238 &             7.14 &               7.23 &                    8.95 &              10.84 \\
    s6 center      &               1,187 &             2.89 &               2.74 &                    4.17 &              29.49 \\
    s6 right       &               1,226 &            12.56 &              12.64 &                   15.04 &              55.90 \\
    \hline
    TOTAL          &              18,398 &             8.59 &               8.45 &                   17.14 &             119.32 \\
    \hline\hline
    \end{tabular}
    \label{tab:fullacc_results}
\end{center}
\end{table*}

\begin{table*}
\begin{center}
    \caption{Evaluation results of the OptScaleVP2 model on the BrnoCompSpeed benchmark}
    \begin{tabular}{|l|r|r|r|r|r|}
    \hline
        video      &     support   &    mean [km/h]   &    median [km/h]   &    95th percentile [km/h]  &     worst [km/h]   \\
    \hline\hline
    s1 left        &                 749 &            17.38 &              17.15 &                   21.18 &              57.54 \\
    s1 center      &                 493 &            48.71 &              47.39 &                   67.73 &             123.94 \\
    s1 right       &                 607 &             5.52 &               5.27 &                    9.38 &              32.51 \\
    s2 left        &               1,030 &            50.70 &              50.25 &                   61.13 &              80.02 \\
    s2 center      &               1,208 &            15.21 &              14.66 &                   19.71 &              38.99 \\
    s2 right       &               1,284 &             4.22 &               4.17 &                    5.17 &              15.99 \\
    s3 left        &                 173 &            15.38 &              15.07 &                   21.15 &              29.44 \\
    s3 center      &                 187 &            12.41 &              12.40 &                   15.64 &              19.60 \\
    s3 right       &                 177 &            27.02 &              26.38 &                   35.72 &              48.87 \\
    s4 left        &               1,026 &            22.05 &              21.92 &                   26.62 &              30.37 \\
    s4 center      &               1,056 &             2.36 &               2.25 &                    2.93 &              26.84 \\
    s4 right       &                 931 &             2.44 &               2.33 &                    3.57 &              20.81 \\
    s5 left        &               1,944 &            12.79 &              12.85 &                   15.45 &              19.80 \\
    s5 center      &               1,932 &             9.31 &               8.79 &                   13.96 &             118.58 \\
    s5 right       &               1,950 &            20.03 &              19.67 &                   29.63 &              81.08 \\
    s6 left        &               1,238 &             6.43 &               6.53 &                    8.13 &               9.94 \\
    s6 center      &               1,187 &            23.87 &              23.99 &                   28.15 &              34.22 \\
    s6 right       &               1,226 &            13.33 &              13.41 &                   15.92 &              56.93 \\
    \hline
    TOTAL          &              18,398 &            15.66 &              13.09 &                   47.86 &             123.94 \\
    \hline
    \end{tabular}
    \label{tab:optscale_results}
\end{center}
\end{table*}

\begin{table*}
\begin{center}
    \caption{Evaluation results of \systemname{} on the BrnoCompSpeed benchmark}
    \begin{tabular}{|l|r|r|r|r|r|}
    \hline
       video      &    support    &    mean [km/h]   &    median [km/h]   &    95th percentile [km/h]  &     worst [km/h]   \\
    \hline\hline
    s1 left       &                 655 &            13.14 &               9.11 &                   39.62 &             114.72 \\
    s1 center     &                 611 &            10.14 &               9.07 &                   23.88 &              53.97 \\
    s1 right      &                 612 &             9.88 &               7.18 &                   28.22 &             173.00 \\
    s2 left       &                 964 &            15.55 &              13.82 &                   23.03 &              78.70 \\
    s2 center     &               1,063 &            40.75 &              43.74 &                   58.30 &              68.84 \\
    s2 right      &               1,145 &            23.81 &              21.26 &                   36.84 &             108.66 \\
    s3 left       &                 134 &             8.95 &               8.81 &                   15.88 &              35.27 \\
    s3 center     &                 147 &             6.43 &               3.95 &                   19.11 &              61.11 \\
    s3 right      &                 140 &            10.98 &              11.37 &                   18.06 &              48.85 \\
    s4 left       &                 883 &            27.01 &              28.26 &                   39.35 &              85.11 \\
    s4 center     &                 831 &            20.06 &              18.21 &                   37.54 &              68.44 \\
    s4 right      &                 860 &            27.04 &              25.63 &                   48.76 &             273.28 \\
    s5 left       &               1,434 &            21.62 &              22.50 &                   36.83 &             383.44 \\
    s5 center     &               1,486 &             7.53 &               4.29 &                   23.96 &             132.71 \\
    s5 right      &               1,491 &            13.70 &              12.63 &                   21.71 &             117.91 \\
    s6 left       &                 965 &            32.46 &              32.35 &                   49.08 &             105.13 \\
    s6 center     &                 929 &            40.10 &              39.87 &                   52.85 &              65.13 \\
    s6 right      &               1,178 &            21.01 &              22.81 &                   29.66 &              64.79 \\
    \hline
    TOTAL         &              15,528 &            21.24 &              19.46 &                   48.18 &             383.44 \\
    \hline
    \end{tabular}
    \label{tab:farsec_results} 
\end{center}
\end{table*}

\begin{table*}
\begin{center}
    \caption{Evaluation results of \systemname{} with blurred input data on the BrnoCompSpeed benchmark}
    \begin{tabular}{|l|r|r|r|r|r|}
    \hline
        video      &     support   &    mean [km/h]   &    median [km/h]   &    95th percentile [km/h]  &     worst [km/h]   \\
    \hline\hline
    s1 left        &                 656 &            11.25 &               7.47 &                   32.78 &             127.96 \\
    s1 center      &                 641 &            13.28 &              11.71 &                   30.55 &              74.47 \\
    s1 right       &                 611 &            11.62 &               9.13 &                   30.72 &             118.60 \\
    s2 left        &                 972 &            14.39 &              13.34 &                   23.05 &              73.39 \\
    s2 center      &               1,057 &            42.77 &              44.38 &                   58.96 &              68.78 \\
    s2 right       &               1,146 &            23.71 &              21.59 &                   36.96 &              84.07 \\
    s3 left        &                 107 &            10.21 &              10.35 &                   16.16 &              18.63 \\
    s3 center      &                 100 &             6.03 &               4.83 &                   17.03 &              22.67 \\
    s3 right       &                 153 &            12.50 &              12.87 &                   21.46 &             104.42 \\
    s4 left        &                 856 &            25.96 &              27.39 &                   37.27 &              75.88 \\
    s4 center      &                 846 &            22.80 &              16.64 &                   47.64 &              85.31 \\
    s4 right       &                 836 &            19.56 &              16.59 &                   40.93 &             332.34 \\
    s5 left        &               1,485 &            17.12 &              18.20 &                   33.06 &             206.68 \\
    s5 center      &               1,459 &             5.48 &               4.15 &                   14.47 &              46.03 \\
    s5 right       &               1,481 &            13.31 &              11.98 &                   22.80 &             135.23 \\
    s6 left        &               1,116 &            27.05 &              26.64 &                   39.16 &              97.45 \\
    s6 center      &               1,114 &            35.16 &              35.30 &                   44.79 &              91.95 \\
    s6 right       &               1,188 &            21.58 &              23.33 &                   29.65 &              72.78 \\
    \hline
    TOTAL          &              15,824 &            20.16 &              18.45 &                   44.93 &             332.34 \\
    \hline
    \end{tabular}
    \label{tab:farsec_blur_results}
\end{center}
\end{table*}

\begin{table*}
    \begin{center}
    \caption{Evaluation results of \systemname{} with noisy input data on the BrnoCompSpeed benchmark}
    \begin{tabular}{|l|r|r|r|r|r|}
    \hline
       video      &     support         &    mean [km/h]   &    median [km/h]   &    95th percentile [km/h]  &     worst [km/h]   \\
    \hline\hline
    s1 left       &                 700 &            23.13 &              19.84 &                   54.91 &             186.54 \\
    s1 center     &                 243 &            36.61 &              30.43 &                   94.10 &             149.06 \\
    s1 right      &                 655 &            26.71 &              21.94 &                   68.50 &             188.03 \\
    s2 left       &                 962 &            17.71 &              14.72 &                   47.87 &             225.86 \\
    s2 center     &                 898 &            28.93 &              30.19 &                   58.50 &             112.30 \\
    s2 right      &               1,083 &            12.45 &               9.93 &                   28.15 &             115.41 \\
    s3 left       &                 125 &             9.59 &               7.62 &                   23.20 &              60.42 \\
    s3 center     &                 104 &             7.59 &               6.91 &                   14.01 &              36.90 \\
    s3 right      &                 131 &            14.20 &               7.48 &                   54.96 &             122.27 \\
    s4 left       &                 874 &            14.62 &              12.58 &                   26.46 &             427.11 \\
    s4 center     &                 793 &            22.06 &              21.91 &                   39.31 &              85.81 \\
    s4 right      &                 862 &            20.52 &              13.84 &                   58.56 &             313.50 \\
    s5 left       &               1,530 &            24.93 &              11.56 &                   91.90 &             338.81 \\
    s5 center     &                 691 &            37.70 &              34.72 &                   81.67 &             181.80 \\
    s5 right      &               1,419 &            34.30 &              27.89 &                  101.46 &             182.72 \\
    s6 left       &               1,048 &            15.02 &              10.40 &                   64.16 &             114.78 \\
    s6 center     &               1,093 &            12.81 &               9.75 &                   29.24 &             167.39 \\
    s6 right      &               1,119 &            10.25 &               8.14 &                   23.49 &              73.61 \\
    \hline
    TOTAL         &              14,330 &            21.28 &              14.53 &                   62.28 &             427.11 \\
    \hline
    \end{tabular}
    \label{tab:farsec_noise_results}
    \end{center}
\end{table*}

\begin{table*}
    \begin{center}
        \caption{Evaluation results of \systemname{} with preprocessing of background removal and fine-tuned YOLO model for this, on the BrnoCompSpeed benchmark}
        \begin{tabular}{|l|r|r|r|r|r|}
            \hline
            video        &   support    &   mean [km/h]   &    median [km/h]   &    95th percentile [km/h]    &   worst [km/h]   \\
            \hline \hline
            s1 left      &                 48 &           27.56 &              17.83 &                     86.86 &           135.67 \\
            s1 center    &                 57 &           16.81 &              12.04 &                     42.23 &            57.50 \\
            s1 right     &                 46 &           13.74 &              11.22 &                     30.98 &            49.79 \\
            s2 left      &                109 &           15.44 &              12.91 &                     38.33 &           103.34 \\
            s2 center    &                133 &           37.04 &              38.46 &                     69.62 &            88.29 \\
            s2 right     &                156 &           22.02 &              19.29 &                     47.96 &           113.83 \\
            s3 left      &                  5 &           10.35 &              10.81 &                     17.68 &            18.44 \\
            s3 center    &                  3 &           24.33 &              12.11 &                     49.91 &            54.11 \\
            s3 right     &                  4 &           16.73 &              13.32 &                     32.64 &            35.67 \\
            s4 left      &                 87 &           26.17 &              25.17 &                     53.21 &            72.48 \\
            s4 center    &                 86 &           22.00 &              18.22 &                     48.09 &            90.80 \\
            s4 right     &                 77 &           31.40 &              27.99 &                     64.75 &            86.12 \\
            s5 left      &                179 &           32.80 &              17.32 &                    130.40 &           280.87 \\
            s5 center    &                284 &           15.30 &              10.27 &                     43.49 &           246.62 \\
            s5 right     &                274 &           16.61 &              11.64 &                     41.57 &           138.22 \\
            s6 left      &                288 &           31.87 &              32.43 &                     44.52 &           105.19 \\
            s6 center    &              1,120 &           35.90 &              37.75 &                     48.39 &            80.93 \\
            s6 right     &                133 &           19.91 &              19.23 &                     37.15 &            51.05 \\
            \hline
            TOTAL        &              3089 &           28.02 &              29.20 &                     50.54 &           280.87 \\
            \hline
        \end{tabular}
        \label{tab:farsec_bg_results}
\end{center}
\end{table*}

\begin{table}
    \begin{center}
        \caption{Evaluation results of \systemname{} applied on the CCTV benchmark \cite{robo_auto_mono} (videos, for which the vehicle detection fails, are excluded).}
        \begin{tabular}{|c|r|r|}
            \hline
             video\_id  &   abs\_errors  &   rel\_errors \\
            \hline \hline
                     1 &      2.43 &     4.48 \\
                     8 &     15.47 &    36.8 \\
                     9 &     19.98 &    34.63 \\
                    11 &      4.78 &    11.81 \\
                    12 &     25.15 &   255.76 \\
                    13 &     29.19 &    75.44 \\
                    14 &     21.80 &    57.34 \\
                    15 &     16.35 &    36.28 \\
                    16 &     13.53 &    29.34 \\
                    20 &     13.49 &    79.03 \\
                    21 &     11.19 &    38.56 \\
                    24 &     19.38 &    40.82 \\
                    25 &      5.58 &    12.79 \\
                    32 &     10.06 &    36.38 \\
                    33 &      2.98 &     7.42 \\
                    34 &      6.25 &    12.73 \\
                    35 &     22.11 &   114.49 \\
                    37 &      9.88 &    36.11 \\
                    38 &      4.76 &    10.83 \\
                    39 &      0.3 &     0.80 \\
                    40 &      1.93 &     4.58 \\
            \hline
        \end{tabular}
        \label{tab:cctv_full_results}
\end{center}
\end{table}

\section{Further Examples and Edge Cases}

\begin{figure}[tb]
    \centering
    \includegraphics[width=0.8\columnwidth]{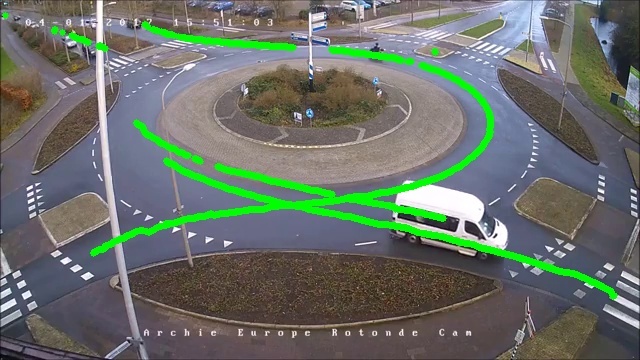}\\
    \vspace*{0.5mm}
    \includegraphics[width=0.8\columnwidth]{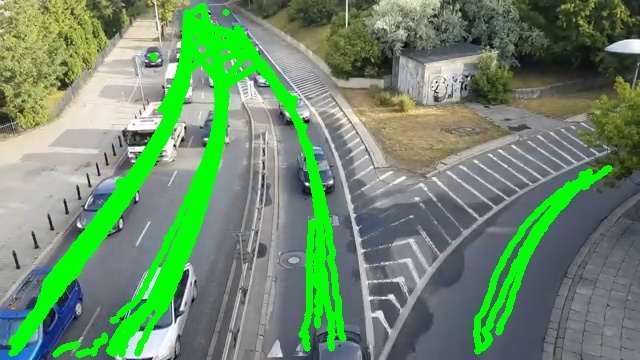}
    \caption{Two example frames from example YouTube datasets with \systemname{} tracked bounding boxes center points visualized. Upper frame: from a CCTV of Regiopurmerend.nl \cite{video_roundabout} displaying a roundabout. Lower frame: from Karol Majek showing curved roads
    \cite{video_curvedroad}.}
    \label{fig:round}
\end{figure}

\begin{figure}[tb]
    \centering
    \includegraphics[width=0.8\columnwidth]{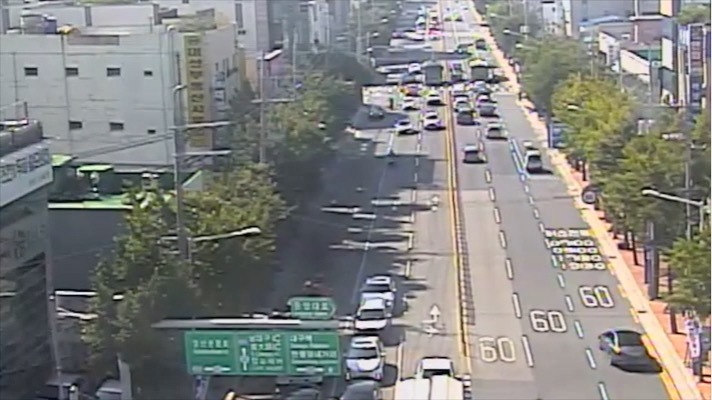}
    \includegraphics[width=0.8\columnwidth]{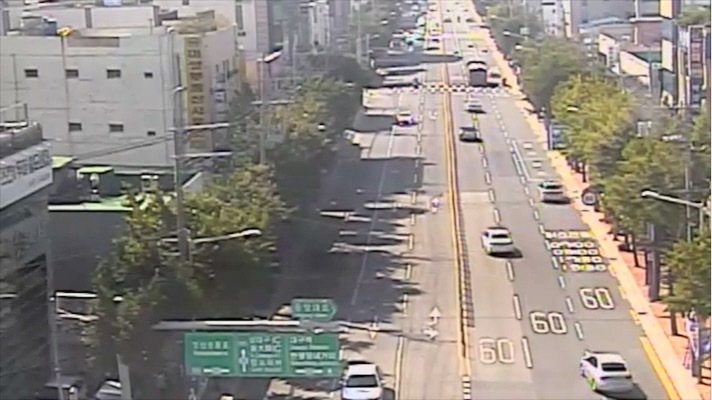}
    \caption{Two example frames from the CCTV dataset \cite{robo_auto_mono}, where our proposed pipeline falsely does not detect vehicles (video-IDs 28 -- top and 29 -- bottom)}
    \label{fig:frames_cctv}
\end{figure}

In our experiments with the CCTV dataset provided by Revaud \etal \cite{robo_auto_mono}, the OpenTrafficCamMap dataset \cite{welch2020open}, and additional traffic camera streams available online \cite{video_curvedroad, video_roundabout}, we have seen that our object detection model handles some edge cases very well, while facing difficulties with others.
Interesting edge cases include curvy roads and a challenging sub-case of them, roundabouts.
While the example video shown in Figure \ref{fig:round} is short and starts when cars are already in the roundabout, the model shows good performance in consistently tracking the vehicles (as indicated by the green lines), which include cars and a van.
Objects that break the line of sight to the vehicle, in this case a pole, interrupt the tracking for a short period (which leads to the assignment of a new ID to the recovered vehicle). 
However, this is not an issue for our approach, in which we calculate the average speed of all vehicles over a sliding window.
Figure \ref{fig:frames_cctv} shows two frames in which our YOLO model does not track vehicles correctly. 
We have identified two reasons for this behavior: 
The vehicles appear very small in those exact frames due to the height of the camera with respect to the street, and the camera feed is flickering in the respective videos, which appears to confuse the YOLO model (video IDs 28 and 29). 
This could be addressed in future work by replacing the YOLOv4 model with a more sophisticated object tracking model, or implementing a method to reassign the previous ID to recovered vehicles.
Based on our analysis of the other baseline models, we assume that the other models experience the same issue.

\begin{figure}[htp]
    \centering
    \includegraphics[width=0.8\columnwidth]{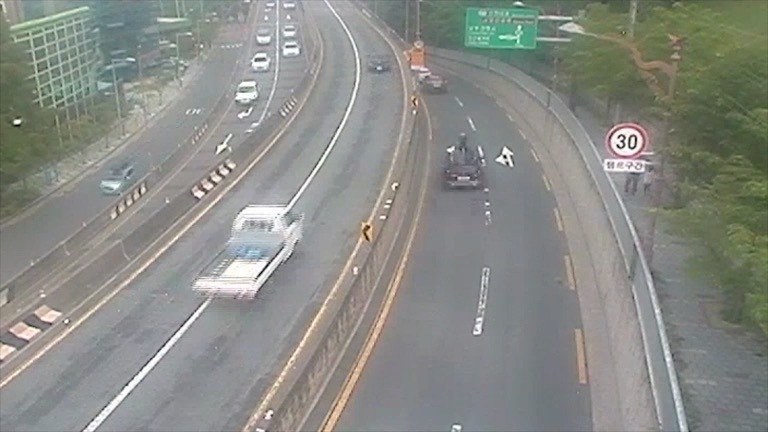}
    \caption{Example frame from the CCTV dataset \cite{robo_auto_mono}, where our proposed pipeline provides an inaccurate average vehicle speed estimation, as multiple streets with the same driving direction but different speeds are visible (video-ID: 22).}
    \label{fig:more_streets}
\end{figure}

In video ID 22 of the CCTV dataset (as shown in Figure \ref{fig:more_streets}), there are four different street segments at different heights included. 
Since \systemname{} currently only distinguishes between driving directions and calculates average speed estimates for either vehicles driving \emph{away} or \emph{towards} the camera, result accuracy will be subpar in these rather complex scenes. 
The example in Figure \ref{fig:more_streets} shows lanes that are part of a highway and others that are not. 
In this specific case, the vehicle speed on the different lanes differs significantly (ranging between standing still and regular highway driving speed) and will currently result in wrong speed estimates. 
In future work, a modular addition for lane detection could fix this issue -- or alternatively, the vehicle speed estimation could be done per vehicle instead of a sliding window for all vehicles per driving direction.

\section{The OpenTrafficCamMap Dataset}

The OpenTrafficCamMap dataset \cite{welch2020open} is a crowd-sourced database of 32,221 traffic cameras located across the United States of America.
The videos are encoded in varying standards, including H.264, VP8, VP9, and JPEG.
Characteristics of the OpenTrafficCamMap data are given in Table \ref{tab:OTCM_table}. Some edge cases found in real-word datasets can be seen in Figure \ref{fig:cctv_challenges}.
Many of the shown real-life challenges are not available in published, recorded benchmark datasets. 
Thus, existing baseline models might deviate significantly from the performance presented in their respective papers.
However, \systemname{} performs well in multiple of these situations and successfully tracks at least a part of the visible vehicles.

Note that support for image streams is only possible with a sufficiently high frame rate (15+ FPS), while most JPEG streams only provide an image once per minute (or less) -- which is why they are marked as partially not supported. 
Some cameras are included in the OpenTrafficCamMap dataset as custom IDs -- these are aggregated as \emph{Others}. 
These could be supported if streams can be provided through the HLS or RTP stream protocols. 
Image or video streams that had an invalid URL are counted as \emph{Invalid}. 
URLs for which the connection or download failed are placed in the category \emph{Error}. 
The remaining URLs that could be successfully accessed are counted as \emph{Available}. 
It should also be considered that multiple sites mentioned in the dataset as image streams also offer M3U8 video streams in non-permanent links. 
An example for this is \emph{fl511.com} with 3,427 image stream entries, while also currently offering over 3,800 video streams, which are not part of the dataset.
Other government-run sites such as \emph{wv511.org} are not mentioned at all for our dataset, while providing many more video streams. 
In addition to the data sources covering the USA, there are further HLS video streams available in the open web. \systemname{} also supports YouTube streams, which are available in multiple different locations. 
Sites such as \emph{webcamtaxi.com} and \emph{camstreamer.com} offer over 900 links to such streams labeled as \emph{traffic}.

\begin{table*}[htp]
    \begin{center}
    \caption{Camera streams available in the OpenTrafficCamMap dataset.}
    \begin{tabular}{|l | r r r r r|}
    \hline
    Stream Format & $\systemname$ compatible & Total & Available & Error & Invalid \\
    \hline
    \hline
    RTP & \checkmark & 0 & - & - & - \\
    \hline
    M3U8 (HLS) & \checkmark & 6,855 & 5,074 & 1,780 & 1 \\
    \hline
    Image & (x) & 23,667 & 14,678 & 8,987 & 2 \\
    \hline
    Others & (x) & 1,699 & - & - & - \\
    \hline
    \end{tabular}
    \label{tab:OTCM_table}
    \end{center}
\end{table*}

\newcommand{\figgrid}{0.3}
\newcommand{\figspace}{\hspace{1em}}
\begin{figure*}
    \begin{subfigure}[b]{0.3\textwidth}
        \includegraphics[width=\columnwidth]{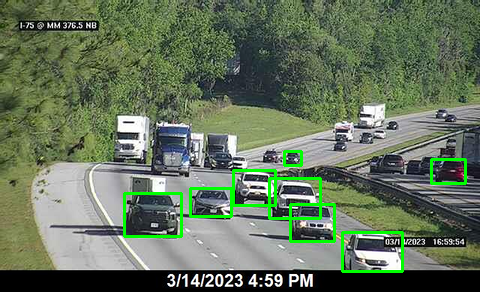}
        \caption{High-distance street view with curve and height differences, lanes on the right only shown partially}
    \end{subfigure}
    \figspace
    \begin{subfigure}[b]{0.3\textwidth}
        \includegraphics[width=\columnwidth]{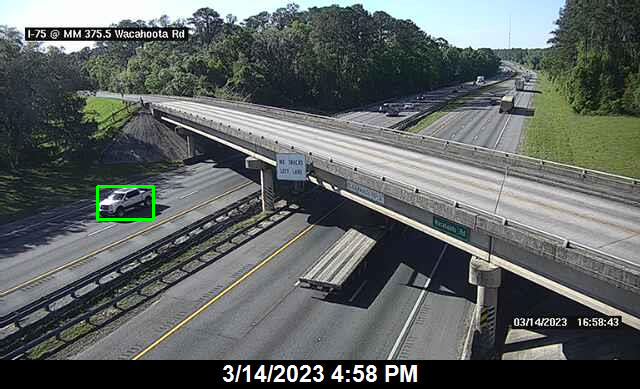}
        \caption{Bridge obstructs view on highway, cars passing over the bridge occasionally in a third direction}
    \end{subfigure}
    \figspace
    \begin{subfigure}[b]{0.3\textwidth}
        \includegraphics[width=\columnwidth]{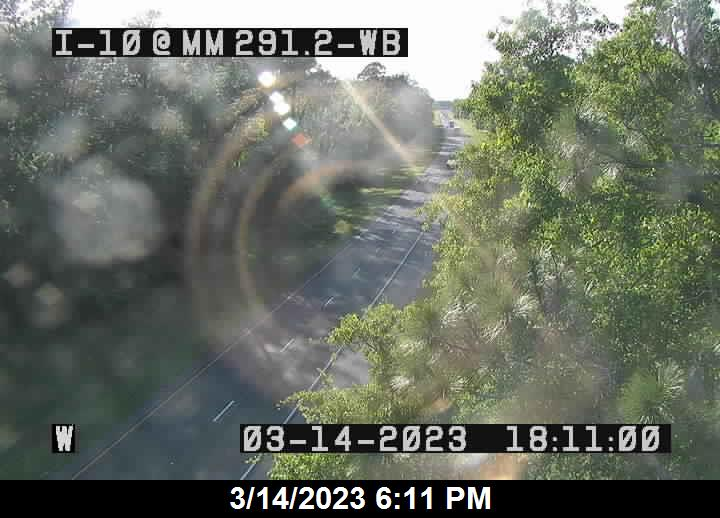}
        \caption{A dirty lens together with lens flare caused by direct sunlight, viewing a country road}
    \end{subfigure}
    \\[1em]
    \begin{subfigure}[b]{0.3\textwidth}
        \includegraphics[width=\columnwidth]{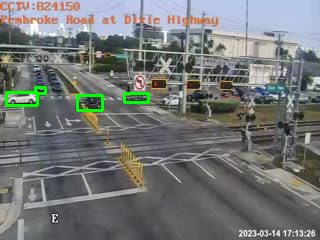}
        \caption{Urban intersection with railway crossing and reduced screen space for the actual vehicle traffic}
    \end{subfigure}
    \figspace
    \begin{subfigure}[b]{0.3\textwidth}
        \includegraphics[width=\columnwidth]{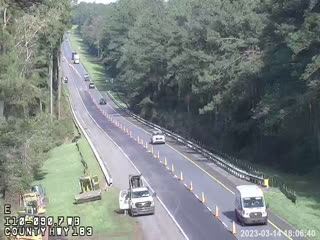}
        \caption{Road construction with a closed lane, affecting traffic routing and conditions}
    \end{subfigure}
    \figspace
    \begin{subfigure}[b]{0.3\textwidth}
        \includegraphics[width=\columnwidth]{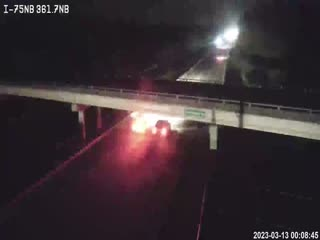}
        \caption{Night-time footage with a bridge blocking view on the highway, headlights blurred due low-light insensitive camera}
    \end{subfigure}
    \\[1em]
    \begin{subfigure}[b]{0.3\textwidth}
        \includegraphics[width=\columnwidth]{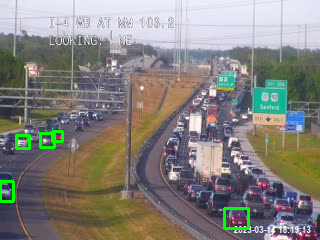}
        \caption{Very dense traffic during rush hour leading to overlapping cars at slow speed, and a cutoff road on the left}
    \end{subfigure}
    \figspace
    \begin{subfigure}[b]{0.3\textwidth}
        \includegraphics[width=\columnwidth]{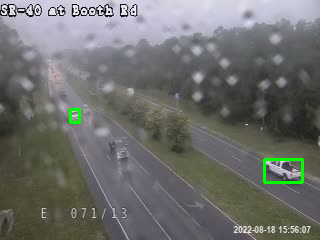}
        \caption{Raindrops on the camera lens blur the image and impede clarity of vision and distorting the perception of vehicles}
    \end{subfigure}
    \figspace
    \begin{subfigure}[b]{0.3\textwidth}
        \includegraphics[width=\columnwidth]{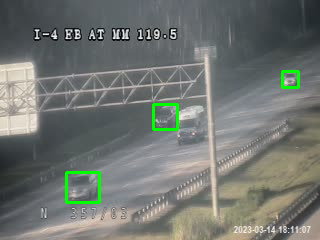}
        \caption{Tilted camera with limited viewing angle and a sign with pole obstructing the view onto the highway}
    \end{subfigure}
    \\[1em]
    \begin{subfigure}[b]{0.3\textwidth}
        \includegraphics[width=\columnwidth]{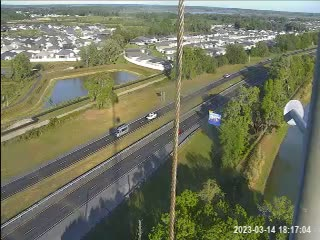}
        \caption{Very high point of view, causing multiple challenges for car tracking}
    \end{subfigure}
    \figspace
    \begin{subfigure}[b]{0.3\textwidth}
        \includegraphics[width=\columnwidth]{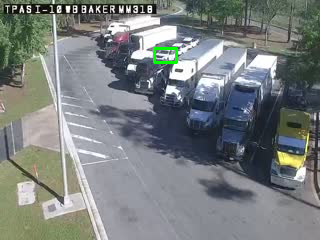}
        \caption{Parked trucks in a highway service area, overlapping each other}
    \end{subfigure}
    \figspace
    \begin{subfigure}[b]{0.3\textwidth}
        \includegraphics[width=\columnwidth]{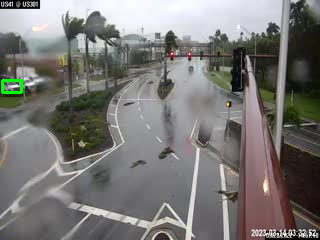}
        \caption{Rain drops on the camera leans distorting or blocking parts of the image}
    \end{subfigure}
    \caption{A variety of scenarios encountered in FL511 traffic cameras from the OTCM dataset \cite{welch2020open}. Vehicles tracked successfully by \systemname{} indicated with green rectangles.}
    \label{fig:cctv_challenges}
\end{figure*}

\section{\systemname{} Pipeline Variants}

\begin{figure}[tb]
    \centering
    \includegraphics[width=0.8\columnwidth]{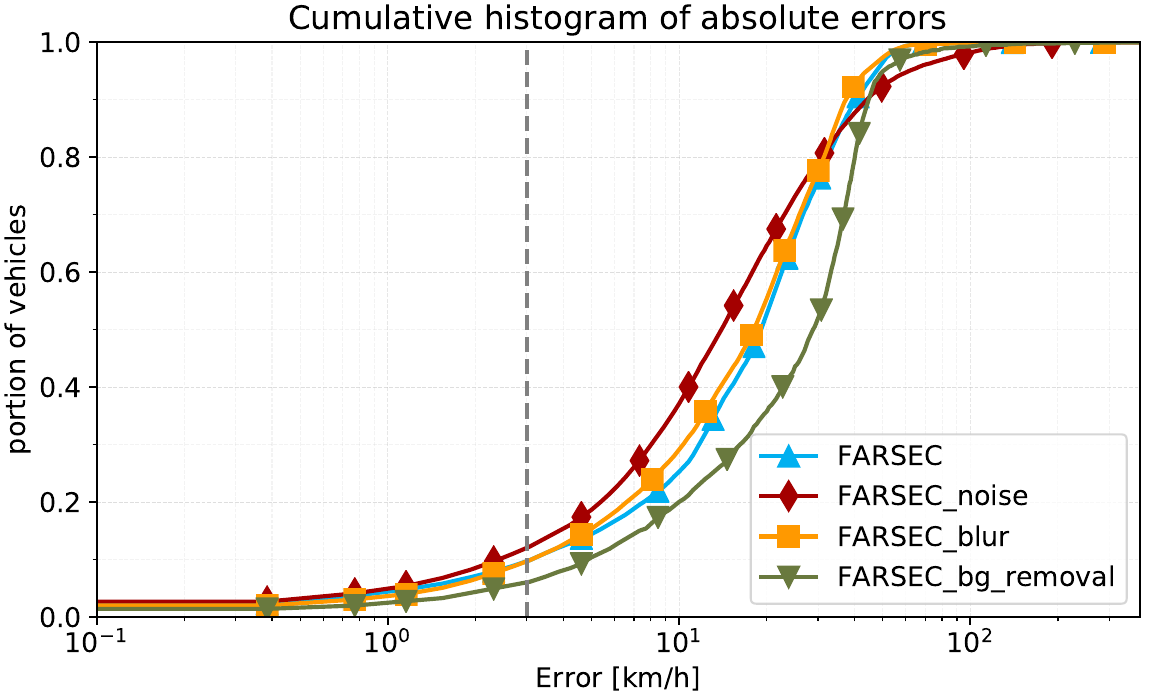}
    \caption{The cumulative absolute error of the different \systemname{} variants on BrnoCompSpeed \cite{compr_brno} plotted in a histogram, including our model's benchmarking results for the augmented datasets with noise and a blur filter and the results with a custom YOLO model finetuned preprocessed and finetuned for background removed frames. The threshold of a 3 km/h error is indicated by the gray vertical dashed line.}
    \label{fig:diff_tracker_cummul}
\end{figure}

The inherent modularity of \systemname{} opens up the possibility of replacing parts of the pipeline and optimizing its speed estimation to cater to particular use cases. 
In our experiments, we included an alternative technique for vehicle tracking based on a pretrained YOLOv4 model, fine-tuned by us on approx.~1,300 manually labelled images with their background removed. In this case, the pipeline includes a preprocessing step for background removal drawing on OpenCV. Only then does our fine-tuned YOLO model detect and track vehicles. The results of using this experimental vehicle detection in comparison to our original \systemname{} option can be seen in Figure \ref{fig:diff_tracker_cummul} and Table \ref{tab:all_approaches_best_results}. 
As shown in the results, this detection method results in a higher mean error on the BrnoCompSpeed dataset and is not used in our final version of \systemname{}. 

During development, we have experimented with alternative depth estimation models, particularly MiDaS \cite{ranftl2020towards}. 
However, this yielded significantly worse results compared to the Pixelformer model \cite{agarwal2023attention}, which we are using in the final \systemname{} pipeline.

\begin{table*}[tb]
    \begin{center}
        \caption{Evaluation results showing mean absolute error (km/h) of all approaches on the BrnoCompSpeed dataset}
        \begin{tabular}{|l|r|r|r|r|r|r|}
        \hline
            Video & FullACC & OptScaleVP2 & \systemname{} & \systemname{} blur & \systemname{} noise & \systemname{} bg removal \\
        \hline \hline
            s1 center & 10.15 & 48.71 & \textbf{10.14} & 13.28 & 36.61 & 16.81 \\ 
            s1 left & \textbf{11.18} & 17.38 & 13.14 & 11.25 & 23.13 & 27.56 \\ 
            s1 right & 6.81 & \textbf{5.52} & 9.88 & 11.62 & 26.71 & 13.74 \\ 
            s2 center & \textbf{11.14} & 15.21 & 40.75 & 42.77 & 28.93 & 37.04 \\ 
            s2 left & 16.24 & 50.70 & 15.55 & \textbf{14.39} & 17.71 & 15.44 \\ 
            s2 right & \textbf{0.95} & 4.22 & 23.81 & 23.71 & 12.45 & 22.02 \\ 
            s3 center & 13.96 & 12.41 & 6.43 & \textbf{6.03} & 7.59 & 24.33 \\ 
            s3 left & 14.28 & 15.38 & \textbf{8.95} & 10.21 & 9.59 & 10.35 \\ 
            s3 right & \textbf{7.54} & 27.02 & 10.98 & 12.50 & 14.20 & 16.73 \\ 
            s4 center & \textbf{1.10} & 2.36 & 20.06 & 22.80 & 22.06 & 22.00 \\ 
            s4 left & 15.88 & 22.05 & 27.01 & 25.96 & \textbf{14.62} & 26.17 \\ 
            s4 right & \textbf{2.05} & 2.44 & 27.04 & 19.56 & 20.52 & 31.40 \\ 
            s5 center & 9.64 & 9.31 & 7.53 & \textbf{5.48} & 37.70 & 15.30 \\ 
            s5 left & \textbf{6.99} & 12.79 & 21.62 & 17.12 & 24.93 & 32.80 \\ 
            s5 right & \textbf{12.02} & 20.03 & 13.70 & 13.31 & 34.30 & 16.61 \\ 
            s6 center & \textbf{2.89} & 23.87 & 40.10 & 35.16 & 12.81 & 35.90 \\ 
            s6 left & 7.14 & \textbf{6.43} & 32.46 & 27.05 & 15.02 & 31.87 \\ 
            s6 right & 12.56 & 13.33 & 21.01 & 21.58 & \textbf{10.25} & 19.91 \\
            \hline
            TOTAL & \textbf{8.59} & 15.66 & 21.24 & 20.16 & 21.28 & 28.02 \\ 
            \hline
        \end{tabular}
    \label{tab:all_approaches_best_results}
    \end{center}
\end{table*}

\end{document}